\documentclass[10pt,twocolumn,letterpaper]{article}

\usepackage{cvpr}              

\AtBeginDocument{%
  \Crefname{subsection}{Subsection}{Subsections}%
  \crefname{subsection}{Subsec.}{Subsecs.}%
}

\definecolor{cvprblue}{rgb}{0.21,0.49,0.74}
\usepackage[pagebackref,breaklinks,colorlinks,allcolors=cvprblue]{hyperref}

\usepackage{amsmath,amssymb,amsfonts}

\usepackage[table]{xcolor}

\usepackage{booktabs}   
\usepackage{multirow}   
\usepackage{makecell}   
\usepackage{longtable}  
\usepackage{siunitx}    
\usepackage{float}    
\usepackage{algorithm}
\usepackage{algorithmic}

\usepackage{listings}

\usepackage{rotate}     
\usepackage{placeins}
\usepackage{cuted}
\usepackage{caption}
\def\paperID{*****} 
\def\confName{CVPR}
\def\confYear{2026}

\title{\raisebox{-3pt}{\includegraphics[height=19pt]{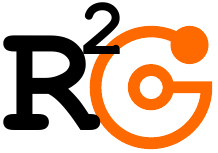}}
: A Multi-View Circuit Graph Benchmark Suite from RTL to GDSII
 \\
\vspace{-8pt}}

\author{
Zewei Zhou$^{1}$,
Jiajun Zou$^{1}$,
Jiajia Zhang$^{1}$,
Ao Yang$^{1}$,
Ruichao He$^{1}$,
Haozheng Zhou$^{1}$\\
Ao Liu$^{1}$,
Jiawei Liu$^{2}$,
Leilei Jin$^{2}$,
Shan Shen$^{1,*}$,
Daying Sun$^{1,*}$ \\
$^{1}$Nanjing University of Science and Technology, $^{2}$The Chinese University of Hong Kong \\
{\tt\small
\{zhouzewei, shanshen, hasdysun\}@njust.edu.cn
}
}
\begin{document}
\maketitle
\begin{abstract}
Graph neural networks (GNNs) are increasingly applied to physical design tasks such as congestion prediction and wirelength estimation, yet progress is hindered by inconsistent circuit representations and the absence of controlled evaluation protocols.
We present \textbf{R2G} (RTL-to-GDSII), a multi-view circuit-graph benchmark suite that standardizes \emph{five stage-aware views} with \emph{information parity} (every view encodes the same attribute set, differing only in where features attach) over 30 open-source IP cores (up to $10^6$ nodes/edges). R2G provides an end-to-end DEF-to-graph pipeline spanning synthesis, placement, and routing stages, together with loaders, unified splits, domain metrics, and reproducible baselines. By decoupling representation choice from model choice, R2G isolates a confound that prior EDA and graph-ML benchmarks leave uncontrolled.
In systematic studies with GINE, GAT, and ResGatedGCN, we find: (i)~view choice dominates model choice, with Test R$^2$ varying by more than 0.3 across representations for a fixed GNN; (ii)~node-centric views generalize best across both placement and routing; and (iii)~decoder-head depth (3--4 layers) is the primary accuracy driver, turning divergent training into near-perfect predictions (R$^2$$>$0.99). Code and datasets: \url{https://github.com/ShenShan123/R2G}.
\end{abstract}
\begingroup
\renewcommand\thefootnote{\fnsymbol{footnote}}
\footnotetext[1]{Corresponding authors}
\endgroup
\section{Introduction}

AI for Electronic Design Automation (EDA) is gaining rapid attention. A growing range of physical design tasks, including placement, routing, timing, and congestion prediction, can be tackled by deep learning models \cite{liao2023dreamplace, shen2024deep, yang2023cnn, jiang2023accelerating}. Modern circuit layouts are structured spatial scenes where components occupy two-dimensional coordinates and interact through connectivity relations. Yet circuit graphs carry structural complexity absent from general-purpose graph datasets: the graph ML community has built rich resources for citation networks \cite{giles1998citeseer}, social graphs \cite{snapnets}, molecular property prediction \cite{Ramakrishnan2014QM9, Sterling2015ZINC15}, and biological networks \cite{zitnik2017ohmnet}, and has matured around benchmarks such as OGB and TUDataset \cite{hu2020open, morris2020tudataset, Dwivedi2020BenchmarkingGNNs}, yet these datasets encode neither typed heterogeneous entities and multi-terminal hyperedges nor geometry-aware attributes. Circuit graphs contain entities with explicit types and hierarchy whose coordinates and orientations are tightly coupled with electrical constraints, spanning multiple non-interchangeable physical design (PD) stages. More critically, the same circuit admits multiple legitimate representations in late physical design, including netlist graphs, spatial graphs, and hypergraphs, whose preserved spatial semantics and message-passing paths differ fundamentally.

\begin{figure}[tb]
    \centering
    \includegraphics[width=\linewidth]{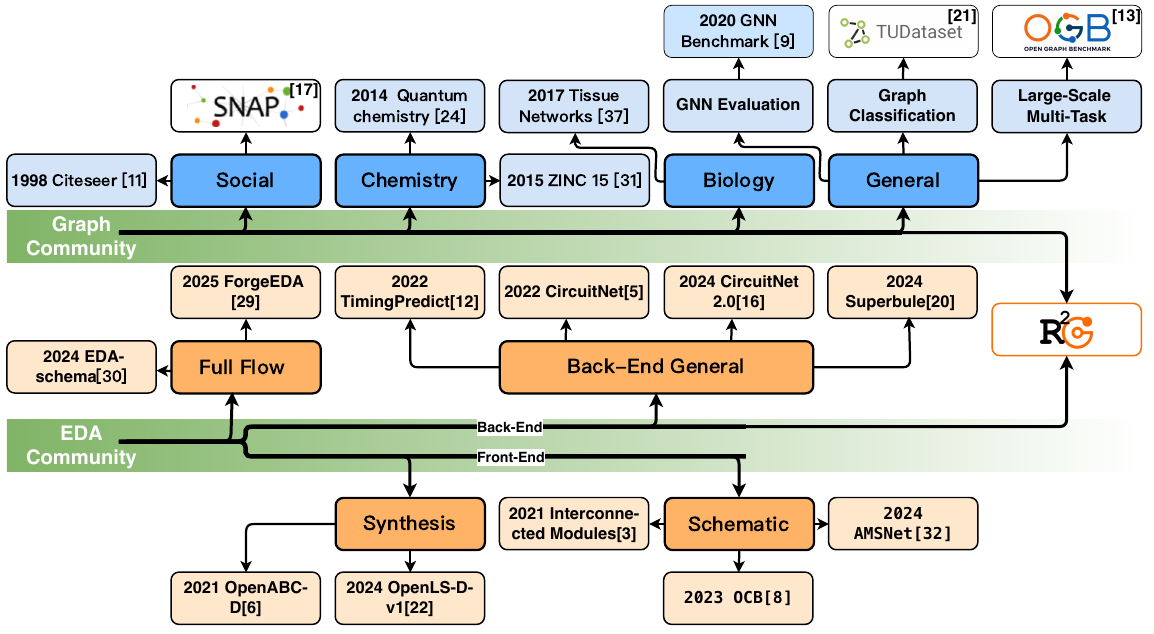}
    \caption{Benchmark and dataset evolution across graph ML and EDA, highlighting R\textsuperscript{2}G’s unique position at their intersection as a stage-aware, multi-view circuit-graph benchmark suite.}
    \label{fig:evolution}
    \vspace{-8pt}
\end{figure}
Despite mature open-source flows and growing graph-oriented resources enabling scalable data generation \cite{openroad_docs, openroad_github}, existing resources cannot support systematic study of graph representation. Front-end datasets are fixed to a single logical representation such as AIG or Boolean circuits, emphasizing logic structure over back-end spatial semantics \cite{chowdhury2021openabc, ni2024openlsdgf}; schematic-level resources focus on netlist understanding and analog topology, remaining distant from digital late-PD tasks \cite{alrahis2022gnnre, dong2023cktgnn, tao2024amsnet}. Back-end datasets co-determine graph representation and task-specific labels, so view choice has never been studied as an independent variable \cite{guo2022timingpredict, chai2022circuitnet, jiang2024circuitnet, luo2024dehnn}. Cross-stage resources have advanced in flow coverage, yet are not designed for the controlled comparison of fixing circuits and tasks while varying only the graph view \cite{shrestha2024edaschema, shi2025forgeeda}. The result is that views, stages, and tasks are deeply entangled, and the independent effect of graph representation on performance remains indiscernible: it is impossible to determine whether a model's accuracy reflects its architecture or its choice of graph representation. \autoref{fig:evolution} situates this gap within the broader evolution of graph ML and EDA datasets.

To address this gap, we propose \textbf{R2G} (\textbf{R}TL-\textbf{to}-\textbf{G}DSII), which standardizes \emph{five stage-aware views} with \emph{information parity} over 30 IP cores, extracting graphs, features, and labels directly from DEF with loaders, unified splits, and reproducible baselines. Our contributions are:

\begin{itemize}

\item \textbf{Multi-view circuit-graph benchmark:} Five complementary views of the \emph{same} late-PD circuits with information parity, stage-aware features, and labels for node-level placement (HPWL) and edge-level routing (wire length), enabling controlled study of representation effects absent from prior single-view EDA datasets.

\item \textbf{Cross-view findings:} For a fixed GNN, Test R$^2$ varies by more than 0.3 across views and model rankings flip, showing that representation choice dominates model selection. Node-centric views are most robust; view~(b) performs best across both stages.

\item \textbf{Decoder-head depth as a dominant lever:} Head depth (3--4 layers) matters far more than GNN depth: raising head layers from 1 to 4 lifts Test R$^2$ from $-$0.17 to 0.99 for placement and resolves NaN divergence for routing~\cite{hu2020open,Dwivedi2020BenchmarkingGNNs}.

\item \textbf{Open-source pipeline:} End-to-end DEF-to-graph pipeline, loaders, unified splits, and baselines over datasets spanning up to $10^6$ nodes/edges. Usage examples: \autoref{sec:usage}.

\end{itemize}

\section{Related Work}
R2G sits at the intersection of graph-ML benchmarking and EDA datasets; we discuss each in turn and then articulate the gap both communities have left open. A detailed comparison table appears in \autoref{sec:graph_datasets}.

\subsection{Benchmarks in the Graph ML Field}
Graph ML benchmarking has matured around large-scale, standardized datasets. OGB~\cite{hu2020open} introduced unified splits and metrics across node, link, and graph tasks, and TUDataset~\cite{morris2020tudataset} consolidated graph classification collections. Datasets for social and citation networks~\cite{snapnets, cora_website, giles1998citeseer}, molecules~\cite{Dwivedi2020BenchmarkingGNNs, Sterling2015ZINC15, Ramakrishnan2014QM9}, and proteins~\cite{zitnik2017ohmnet, hu2020open} established best practices for loaders, baselines, and evaluation. Despite these advances, all of these datasets are domain-agnostic and predominantly single-view: they encode neither typed heterogeneity and multi-terminal hyperedges nor geometry-aware attributes required for physical design, and none offers multiple representations of the \emph{same} data to isolate representation effects.

Circuit graphs differ fundamentally from these benchmarks: a single design reaches $10^4$--$10^6$ nodes and edges in incidence-style views, while geometry- or routing-resource views add $\mathcal{O}(W\!\times\!H\!\times\!L)$ edges (where $W$, $H$, and $L$ denote layout width, height, and metal-layer count); DEF files provide explicit 2D layout cues (cell coordinates, orientations); and both topology and spatial structure are intrinsic to the representation.

\subsection{Benchmarks in the EDA Field}
Open EDA datasets have advanced ML for circuits across stages and tasks. Front-end datasets such as OpenABC~\cite{chowdhury2021openabc} and OpenLS-D-v1~\cite{ni2024openlsdgf} are built on fixed logical representations like AIG or Boolean circuits, emphasizing synthesis over back-end spatial semantics. Schematic-level resources including GNN-RE~\cite{alrahis2022gnnre}, CktGNN~\cite{dong2023cktgnn}, and AMSNet~\cite{tao2024amsnet} address netlist understanding and analog topology outside the digital late-PD scope. Back-end datasets including CircuitNet~\cite{chai2022circuitnet}, CircuitNet2.0~\cite{jiang2024circuitnet}, the TimingPredict dataset~\cite{guo2022timingpredict}, and Superblue~\cite{luo2024dehnn} target congestion, timing, and routability tasks, but co-determine graph representation and labels---so view choice has never been studied as an independent variable. Cross-stage resources such as EDA-schema~\cite{shrestha2024edaschema} and ForgeEDA~\cite{shi2025forgeeda} have advanced flow coverage and multimodal aggregation, alongside EDALearn~\cite{Pan2023EDALearnAC} and synthetic generators~\cite{hutton2002characterization}, yet none is designed for 
controlled cross-view comparison under fixed circuits and tasks.

Yet the field still lacks \emph{standard graph benchmarks}: existing resources adopt a single representation per task or stage with inconsistent node/edge types, features, and label resolutions, and no unified conversion from design files to typed graphs with information parity across stages. The key gap is therefore benchmark formulation rather than data availability. R2G addresses this by inheriting graph-ML best practices (unified splits, scalable loaders, reproducible baselines) while providing typed, multi-view circuit graphs with information parity and resolution-matched supervision, offering the first controlled framework for evaluating circuit graph representations across PD stages.
\section{\raisebox{-2pt}{\includegraphics[height=13pt]{figs/r2g.pdf}} Benchmark Construction}
We now detail how R2G is constructed, beginning with the design corpus and its graph-relevant characteristics.

\subsection{Digital Designs}

\begin{table}[ht]
\centering
\caption{Design corpus statistics including circuit sizes and corresponding graph scale (\#Nodes and \#Edges under view~(b)) for representative open-source IP cores.}
\resizebox{\linewidth}{!}{
\begin{tabular}{lp{0.30\linewidth}cccccp{0.34\linewidth}}
\toprule
\multirow{2}{*}{\textbf{Designs}} & \multirow{2}{*}{\textbf{Category}} & \multicolumn{5}{c}{\textbf{Circuit \& Graph Statistics}} & \multirow{2}{*}{\textbf{Function}} \\
\cmidrule(lr){3-7}
 & & \textbf{\#Gates} & \textbf{\#Nets} & \textbf{\#IOs} & \textbf{\#Nodes} & \textbf{\#Edges} & \\
\midrule

ss\_pcm & Audio controller & 463 & 521 & 28 & 2.44k & 2.88k & pulse code modulation \\
ac97\_ctrl & Audio controller & 10509 & 12531 & 132 & 58.22k & 70.60k & AC97 Audio Codec Controller \\
vga\_lcd & Video controller & 97688 & 110587 & 198 & 531.33k & 650.84k & VGA/LCD Display Controller \\ 
\midrule

des3\_area & Crypto core & 1529 & 1655 & 190 & 9.78k & 12.46k & 3DES Enc/Dec Module \\ 
systemcdes & Crypto core & 2663 & 2788 & 197 & 14.17k & 17.02k & SystemC DES Encryption/Decryption Core \\
systemcaes & Crypto core & 7565 & 7963 & 389 & 41.08k & 50.29k & SystemC AES Encryption/Decryption Core \\
sha256 & Crypto core & 10637 & 12461 & 774 & 62.83k & 76.32k & SHA256 Hash Accelerator \\
aes\_secworks & Crypto core & 21700 & 23941 & 520 & 128.40k & 164.11k & Area-optimized AES Implementation \\
aes\_xcrypt & Crypto core & 28337 & 29595 & 391 & 165.76k & 215.07k & AES Engine with Enc/Dec Support \\ 
\midrule

tv80 & Processor & 5634 & 5880 & 63 & 31.14k & 39.45k & Z80 Compatible Processor Core \\
tv80s & Processor & 7298 & 7377 & 46 & 37.24k & 45.74k & Z80 Compatible Processor Core \\
riscv32i & Processor & 9712 & 10446 & 135 & 53.23k & 66.91k & Simplified RISC Variant Processor \\
ibex & Processor & 16523 & 18760 & 264 & 96.60k & 122.18k & Ibex 32-bit RISC-V CPU Core \\
tinyRocket & Processor & 27770 & 32239 & 269 & 50.63k & 36.27k & TinyRocket RISC-V Processor Core \\
swerv & Processor & 89893 & 102155 & 2039 & 538.36k & 684.89k & SweRV EH1 RISC-V Core \\
bp\_multi & Processor & 158629 & 139704 & 1453 & 296.72k & 250.15k & Black Parrot multi-core processor \\ 
\midrule

uart & Communication controller & 478 & 550 & 54 & 2.73k & 3.29k & Universal Async Receiver/Transmitter \\ 
sasc\_top & Communication controller & 696 & 726 & 28 & 3.45k & 4.17k & Simple Async Serial Comm Controller \\
i2c\_verilog & Communication controller & 883 & 959 & 33 & 4.69k & 5.70k & Inter-Integrated Circuit Bus Controller \\
simple\_spi\_top & Communication controller & 936 & 968 & 28 & 4.57k & 5.49k & Simple SPI Controller \\
spi\_top & Communication controller & 3031 & 3050 & 92 & 15.62k & 19.18k & SPI Controller \\
dynamic\_node & Communication controller & 11196 & 13560 & 693 & 67.87k & 83.87k & Dynamic Network Router Node \\
pci & Communication controller & 12041 & 14915 & 369 & 72.69k & 90.70k & PCI Interface Controller \\
usb\_funct & Communication controller & 12222 & 14121 & 249 & 67.30k & 81.58k & USB Function Controller \\ 
wb\_dma\_top & System controller & 3416 & 3840 & 432 & 19.52k & 23.03k & WISHBONE DMA Controller \\
wb\_conmax & Communication controller & 32161 & 33312 & 2546 & 181.45k & 221.76k & WISHBONE Connection Matrix \\
usb\_phy & Communication controller & 555 & 588 & 33 & 2.73k & 3.22k & USB Physical Layer \\
\midrule

fir & DSP core & 4077 & 4777 & 66 & 22.63k & 27.68k & Finite Impulse Response Filter \\
jpeg & DSP core & 61865 & 73108 & 47 & 331.18k & 394.38k & JPEG Encoder/Decoder \\
idft & DSP core & 138312 & 178202 & 132 & 817.88k & 1009.40k & Inverse Discrete Fourier Transform \\

\bottomrule
\end{tabular}
}
\label{tab:design-statistics}
\end{table}

We begin with the R2G design corpus: 30 open-source IP cores from OpenCores and GitHub spanning processors, DSP, cryptography, communication/system control, and video/audio control. \autoref{tab:design-statistics} summarizes gate and net counts, top-level netlist/DEF IO counts, node and edge counts, and each design's function and category. For discussion, we consolidate these 30 designs into five broad categories and highlight their graph-relevant traits.
\begin{itemize}

  \item \textbf{Video/Audio controllers:} Structured dataflow with modest IO counts.
  
  \item \textbf{Communication/System controllers:} Protocol-centric controllers and bus interconnects emphasize handshakes, arbitration, and high external connectivity; graphs highlight interface nets, control trees, and contention points.

  \item \textbf{Crypto cores:} Dense combinational logic with staged datapaths; regular rounds stress arithmetic-heavy paths.
   
  \item \textbf{DSP cores:} Arithmetic throughput and wide datapaths; structured pipelines and repeated arithmetic motifs.

  \item \textbf{Processors:} Largest circuit and graph scales; diverse fan-in/fan-out; mixed control--datapath structure.
\end{itemize}

Across categories, circuit/graph scale, interface complexity, and topology vary, motivating careful \emph{view selection} and \emph{task formulation}. Compute-dense cores emphasize spatial coordination and placement signals, whereas interface-heavy controllers expose stronger routing, fanout, and congestion patterns.


\subsection{Post-End Flow}
\begin{figure}[tb]
    \centering
    \includegraphics[width=\linewidth]{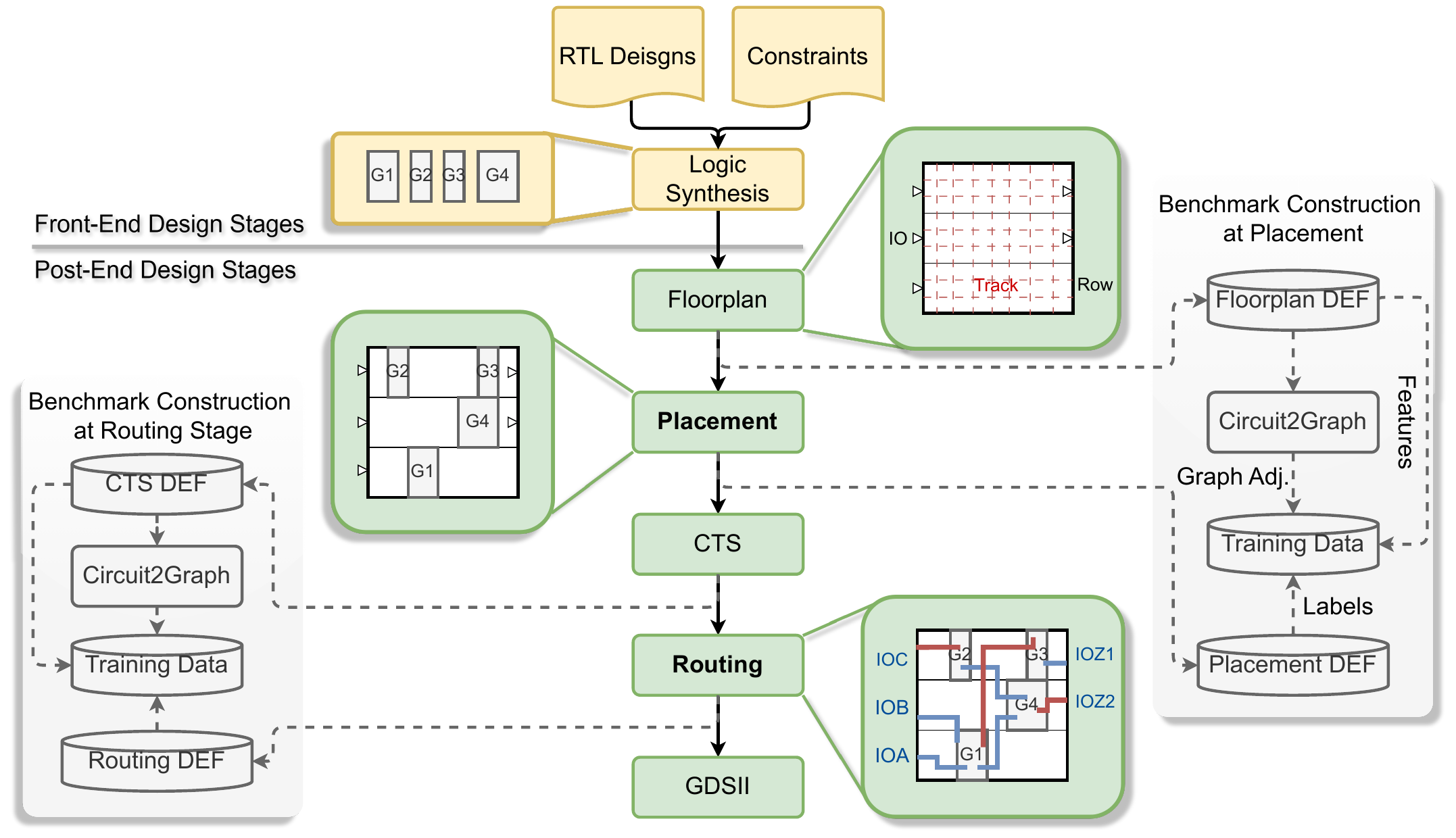}
    \caption{OpenROAD RTL-to-GDSII post-end flow including floorplanning, placement (global and detailed), clock-tree synthesis, routing (global and detailed), and signoff. R2G focuses on placement and routing, while earlier stages provide constraints and contextual information.}
    \label{fig:post-design}
\end{figure}

\begin{figure*}[t]
    \centering
    \includegraphics[width=\textwidth]{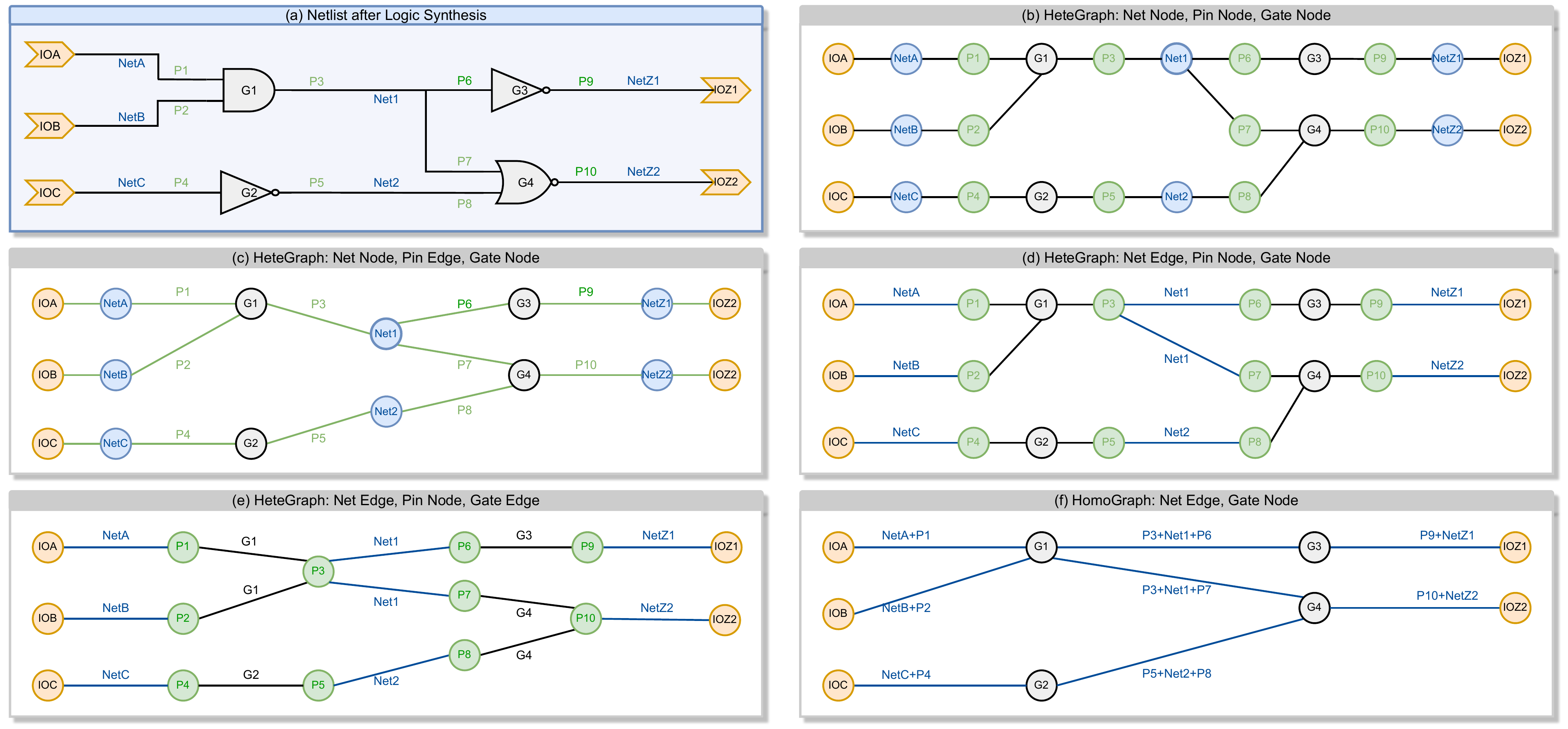}
    \captionof{figure}{Original circuit schematic (a) and five complementary circuit graph views (b--f) used in R2G: (b) all-elements-as-nodes, (c) pins-as-edges, (d) nets-as-edges, (e) net--gate incidence edges, (f) net edges without pin nodes. These views support placement and routing analyses under different supervision granularities.}
    \label{fig:net2g}
\end{figure*}

Physical design transforms RTL into manufacturable layouts through the RTL-to-GDSII post-end flow. Following the OpenROAD pipeline (\autoref{fig:post-design}), the flow includes synthesis, floorplanning, macro placement, global and detailed placement, clock tree synthesis (CTS), global and detailed routing, and signoff verification \cite{openroad_github}. DEF files are produced after floorplanning, placement, and routing, together with routing guides and timing or congestion reports, encoding geometric layout, circuit topology, and electrical information that provide the basis for graph construction. A fragment of the DEF template is shown in \autoref{sec:def}.

R2G focuses on the placement and routing stages because they largely determine timing, power consumption, and routability. Supervision is derived from physically meaningful quantities produced by the flow and is attached to whichever entity instantiates the target in a given graph view. For placement, these quantities are derived from placed cell geometry and net bounding boxes; for routing, they describe realized interconnect properties such as routed wire length and via usage.

Placement quality is evaluated using half-perimeter wirelength (HPWL), derived from bounding boxes of connected gates. Routing quality is measured using the exact routed wirelength after metal layer assignment and via insertion. These labels remain consistent across stages: placement HPWL approximates routing effort, while final routed wirelength reflects realized interconnect across metal layers.

Existing datasets often isolate individual steps such as synthesis trajectories or single-stage layout snapshots~\cite{chowdhury2021openabc, chai2022circuitnet, jiang2024circuitnet, Pan2023EDALearnAC}, leading to fragmented graph representations and evaluation protocols. In contrast, R2G captures the full placement and routing stages and tracks structural changes directly from DEF, including topology updates (e.g., buffering and clock-tree sink insertion) and feature shifts related to placement density, geometric constraints, routing layers, and via usage. Outputs from earlier stages, such as floorplan constraints, CTS sinks, and power grid structures, are incorporated as contextual features rather than prediction targets. This design preserves the modular OpenROAD flow while grounding multi-view circuit graphs in consistent stage-aware supervision.

\subsection{Multi-View Circuit Graphs}

\textbf{Key graph definitions.} We represent \emph{digital} circuits as typed, attributed graphs $G=(\mathcal{V},\mathcal{E})$ whose nodes and edges carry domain features drawn from DEF. A \emph{heterogeneous} circuit graph includes multiple node and edge types to encode EDA semantics faithfully. R2G is built from four fundamental entity types: (i)~\emph{logic gates}, (ii)~directional \emph{pins} (input/output/inout), (iii)~electrical \emph{nets} connecting multiple pins (multi-terminal connectivity), and (iv)~\emph{IOs} (macro ports and pads). Crucially, the same entity may appear as a node or an edge depending on the view: a net becomes a node in bipartite or hypergraph views to capture multi-terminal connectivity, but an edge in pairwise views; pins can similarly be explicit nodes or encoded as incidence attributes on gate--net edges.

We provide an original circuit schematic (a) and five complementary views (b–f) in \autoref{fig:net2g}, each panel tied to a consistent conversion rule and emphasizing distinct logical, geometric, and resource aspects of the same circuit:
\begin{enumerate}
  \item[a)] \textbf{Original circuit schematic:} gate-level functional diagram with signals and IOs prior to graph conversion; serves as the source representation.
  \item[b)] \textbf{All-elements-as-nodes:} gates, pins, nets, and IOs are nodes; typed edges encode relations (connects-to, drives, belongs-to). This maximizes semantic coverage.
  
  \item[c)] \textbf{Pins-as-edges:} pins become typed edges between gate and net nodes; preserves incidence and direction while reducing size.
  
  \item[d)] \textbf{Nets-as-edges:} nets become edges between gate nodes; emphasizes pairwise connectivity.
  
  \item[e)] \textbf{Net–Gate incidence-as-edges:} net--gate membership is encoded with edges in a bipartite formulation; retains multi-terminal semantics and scales to large designs.
  
  \item[f)] \textbf{Net edges without pin nodes:} pin nodes are pruned; nets remain as edges between gates.
\end{enumerate}

\textbf{Task alignment and empirical usage.} Researchers select views to match signals and labels rather than a fixed front-/post-end dichotomy. View (b) supports early-stage coupling capacitance prediction \cite{shen2025few}. View (c) is used for early-stage ground capacitance prediction \cite{shen2024deep} and for placement-stage congestion and \texttt{HPWL} prediction \cite{yang2022versatile}. View (d) emphasizes pairwise connectivity and is adopted for analog circuit topology generation \cite{gao2025analoggenie}. View (e) encodes net–gate membership suitable for timing-graph net delay prediction at the placement stage \cite{jiang2024circuitnet}. View (f) prunes pin nodes for scalability and is widely used for functional and synthesis tasks, including truth table prediction and SAT solving \cite{shi2023deepgate2,shi2024deepgate3}, circuit synthesis \cite{chowdhury2021openabc}, and logic synthesis with PPA prediction \cite{shi2025forgeeda}.

\textbf{View choices across stages.} The appropriate \emph{inductive bias} follows the task: capacitance estimation benefits from direction and membership in view (b/c); placement metrics (congestion, HPWL) and timing rely on geometry attached to gates/edges in view (c/e); analog topology generation leverages pairwise net connectivity in view (d); functional and synthesis tasks favor scalable net-edge view (f). This diversity underscores a fidelity–scalability trade-off and motivates standardized, resolution-matched evaluation across views. More statistical information of five views is compared in \autoref{sec:stat}.

\textbf{R2G standardization.} R2G provides unified DEF-to-graph conversion across five typed, attributed views (b–f), enabling resolution-matched evaluation for placement and routing and controlled cross-view ablations that disentangle connectivity versus geometry signals. Selecting view per task avoids forcing a single representation and supports transfer across designs and technology nodes.

\subsection{Feature \& Label Extraction}\label{sec:feat_label_extraction}

\subsubsection{Feature collection and assembly}
\begin{table}[t]
  \centering
  \caption{Definitions of features/labels for circuit components in terms of views.}
  \label{tab:feat_label_schema}
  \resizebox{\linewidth}{!}{%
  \begin{tabular}{cll p{0.34\linewidth} p{0.4\linewidth}}
    \toprule
    \textbf{Unit}  & \textbf{Name} & \textbf{Type} & \textbf{Definition} & \textbf{Notes} \\
    \midrule
    Gate  & \texttt{x} & feat. (rout.) & Gate x (DBU) & Node (b/c/d/f); edge (e) \\
    Gate  & \texttt{y} & feat. (rout.) & Gate y (DBU) & Node (b/c/d/f); edge (e) \\
    Gate  & \texttt{cell\_type} & feat. & Integer encoding of gate type & Node (b/c/d/f); edge (e) \\
    Gate  & \texttt{orientation} & feat. & Orientation code (0--7) & Node (b/c/d/f); edge (e) \\
    Gate  & \texttt{area} & feat. & Gate area (tech lib) & Node (b/c/d/f); edge (e) \\
    Gate  & \texttt{place\_flag} & feat. & Placement status & Node (b/c/d/f); edge (e) \\
    Gate  & \texttt{power\_leak} & feat. & Leakage power (tech lib) & Node (b/c/d/f); edge (e) \\ \midrule
    Net   & \texttt{net\_type} & feat. & Net category encoding & Node (b/c); edge (d/e/f) \\
    Net   & \texttt{pin\_count} & feat. & Pin count & Node (b/c); edge (d/e/f) \\
    Net   & \texttt{HPWL} & \begin{tabular}[c]{@{}l@{}}label (place.)/\\ feat. (rout.)\end{tabular} & HPWL (DBU) from pin bbox & Node (b/c); edge (d/e/f) \\
    Net   & \texttt{wire\_length} & label (rout.) & Exact routed wire length (DBU), per-layer Manhattan & node (b/c), edge (d/e/f) \\
    Net   & \texttt{via\_count} & label (rout.) & Via count & Node (b/c); edge (d/e/f) \\ \midrule
    IO    & \texttt{x} & feat. & IO x (DBU) & Node (b/c/d/e/f) \\
    IO    & \texttt{y} & feat. & IO y (DBU) & Node (b/c/d/e/f) \\
    IO    & \texttt{orientation} & feat. & IO orientation & Node (b/c/d/e/f) \\
    IO    & \texttt{layer\_id} & feat. & Metal layer index & Node (b/c/d/e/f) \\ \midrule
    Pin   & \texttt{pin\_type} & feat. & Pin type encoding & Node (b); edge (c/e); aggregated (d/f) \\
    Pin   & \texttt{cell\_type} & feat. & Owning gate type encoding & Node (b); edge (c/e); aggregated (d/f) \\
    \bottomrule
  \end{tabular}
  }
\end{table}

We standardize feature definitions for four unit types (gates, nets, IOs, pins) and align them to the multi-view circuit graphs in \autoref{fig:net2g}, views~(b--f). \autoref{tab:feat_label_schema} specifies the schema and sources (DEF and technology libraries). To maintain a coherent narrative, we first describe stage-specific feature sets and then detail how they are attached under each view.

Feature sets are stage-specific. During placement, gates omit \texttt{x}/\texttt{y} and rely on static attributes such as \texttt{cell\_type}, \texttt{area}, and \texttt{power\_leak}. During routing, gates include placed \texttt{x}/\texttt{y}, \texttt{orientation}, and \texttt{place\_flag}. Nets include \texttt{net\_type}, \texttt{pin\_count}, and \texttt{HPWL} as a derived feature once gate positions are determined. IO features comprise \texttt{x}/\texttt{y}, \texttt{orientation}, and \texttt{layer\_id}. Pins expose two categorical entries: \texttt{pin\_type} encodes the pin type, and \texttt{cell\_type} encodes the owning gate type.Coordinates (\texttt{x}, \texttt{y}) are reported in DBU (Database Unit).

To ensure reproducibility and consistent ordering across designs and views, discrete fields use stable integer vocabularies: \texttt{cell\_type} maps logic-gate categories to IDs (e.g., 0--95); \texttt{orientation} maps to codes 0--7; \texttt{net\_type} encodes \{\texttt{signal}, \texttt{power}, \texttt{ground}, \texttt{clock}, \texttt{reset}, \texttt{scan}\}; \texttt{layer\_id} indexes metal layers; and \texttt{place\_flag} encodes placement status.

With features defined, we keep feature parity across views: the same set of attributes is available in every view—no fields are missing and none are duplicated across views. We vary only attachment location to enable fair comparison across \autoref{fig:net2g}, while preserving information content: (b) All-elements-as-nodes keeps gate, pin, net, and IO features on their respective nodes; (c) Pins-as-edges moves pin features to gate--net incidence edges while gates and nets retain node features; (d) Nets-as-edges attaches net features to pairwise gate--gate edges and propagates them deterministically to edges derived from the same net (without introducing new attributes); (e) Net--Gate incidence-as-edges attaches gate features to incidence edges between net and gate nodes to preserve bipartite semantics; and (f) Net edges without pin nodes concatenates pin and net features into a single attribute on gate--gate net edges. When multiple pins exist between a gate pair, pin features are aggregated with a fixed reducer before concatenation to avoid duplicating information. This design ensures feature parity across views and isolates representation effects.

\subsubsection{Label calculation}
We compute task-specific net labels consistently across stages. For placement, supervision uses \texttt{HPWL}, defined as the minimal bounding box over a net’s pins:
\begin{equation}
\mathtt{HPWL} = (x_{\max} - x_{\min}) + (y_{\max} - y_{\min}).
\end{equation}
After placement, \texttt{HPWL} becomes a derived routing feature because gate positions are available.
For routing, supervision uses exact \texttt{wire\_length} from DEF routed geometry. For each metal layer $\ell$, there is
\begin{equation}
\mathtt{wire\_length} = \sum_{\ell \in \mathcal{L}} \sum_{(x_1,y_1),(x_2,y_2)\in \ell} \big|x_2 - x_1\big| + \big|y_2 - y_1\big|.
\end{equation}
Here, $\mathcal{L}$ denotes the routed metal layers parsed from the nets section. We also parse an auxiliary routing label, \texttt{via\_count}, from DEF tokens.

Consistent with the chosen representation, labels attach to the entity that represents the net: when a net is a node (views b/c in \autoref{fig:net2g}), the label attaches to that node; when a net is an edge (views d--f), the label attaches to that edge. As designs progress through physical stages, the supervisory target changes accordingly: \texttt{HPWL} supervises placement and then transitions to a derived feature after placement, whereas \texttt{wire\_length} and \texttt{via\_count} supervise routing. This convention preserves stage-aware supervision while informing downstream views with physically grounded features.

\section{Experiments}

We report three standard regression metrics: mean absolute error (MAE), root mean squared error (RMSE), and coefficient of determination (R\textsuperscript{2}).

To enable fair cross-view comparison, we convert typed, heterogeneous graph views into homogeneous graphs with unified node and edge types while preserving attribute parity across views. This allows classic homogeneous GNNs, GINE~\cite{xu2018powerful}, GAT~\cite{velickovic2018graph}, and ResGatedGCN~\cite{bresson2017residual}, to serve as baselines across all views. In practice, homogeneous GNNs are widely adopted in EDA learning systems for scalability and lower engineering complexity, making them suitable for controlled benchmark comparisons. The experimental setup can be found in \autoref{sec:hp_tune}. We leave heterogeneous representations and hetero-GNNs to future work.

\subsection{Comparison of Different Views for Placement Tasks}

\begin{table*}[t]
\centering
\caption{Placement results across views (b--f) and splits. Lower MAE/RMSE is better; higher R\textsuperscript{2} is better.}
\label{tab:placement_views}
\resizebox{\linewidth}{!}{%
\begin{tabular}{ll| 
    >{\columncolor[HTML]{E0E0E0}}c >{\columncolor[HTML]{E0E0E0}}c >{\columncolor[HTML]{E0E0E0}}c 
    >{\columncolor[HTML]{F9F2E7}}c >{\columncolor[HTML]{F9F2E7}}c >{\columncolor[HTML]{F9F2E7}}c 
    >{\columncolor[HTML]{E6F2FF}}c >{\columncolor[HTML]{E6F2FF}}c >{\columncolor[HTML]{E6F2FF}}c 
    >{\columncolor[HTML]{E6FFE6}}c >{\columncolor[HTML]{E6FFE6}}c >{\columncolor[HTML]{E6FFE6}}c 
    >{\columncolor[HTML]{FFF7E6}}c >{\columncolor[HTML]{FFF7E6}}c >{\columncolor[HTML]{FFF7E6}}c 
    }
\toprule
& & \multicolumn{3}{c}{\cellcolor[HTML]{E0E0E0}\textbf{(b)}} & \multicolumn{3}{c}{\cellcolor[HTML]{F9F2E7}\textbf{(c)}} & \multicolumn{3}{c}{\cellcolor[HTML]{E6F2FF}\textbf{(d)}} & \multicolumn{3}{c}{\cellcolor[HTML]{E6FFE6}\textbf{(e)}} & \multicolumn{3}{c}{\cellcolor[HTML]{FFF7E6}\textbf{(f)}} \\ 
\cmidrule(lr){3-5} \cmidrule(lr){6-8} \cmidrule(lr){9-11} \cmidrule(lr){12-14} \cmidrule(lr){15-17}
\textbf{Split} & \textbf{Model} & \textbf{MAE $\downarrow$} & \textbf{RMSE $\downarrow$} & \textbf{R\textsuperscript{2} $\uparrow$} & \textbf{MAE $\downarrow$} & \textbf{RMSE $\downarrow$} & \textbf{R\textsuperscript{2} $\uparrow$} & \textbf{MAE $\downarrow$} & \textbf{RMSE $\downarrow$} & \textbf{R\textsuperscript{2} $\uparrow$} & \textbf{MAE $\downarrow$} & \textbf{RMSE $\downarrow$} & \textbf{R\textsuperscript{2} $\uparrow$} & \textbf{MAE $\downarrow$} & \textbf{RMSE $\downarrow$} & \textbf{R\textsuperscript{2} $\uparrow$}\\
\midrule
\multirow{3}{*}{Train} 
& GINE & \textbf{0.2859} & \textbf{0.6117} & \textbf{0.9554} & 0.5061 & 0.8678 & 0.9017 & 0.3272 & 0.4516 & 0.7984 & \textbf{0.1649} & \textbf{0.2936} & \textbf{0.9137} & \textbf{0.3334} & \textbf{0.4563} & \textbf{0.7941} \\
& ResGatedGCN & 0.3097 & 0.655 & 0.9489 & \textbf{0.4808} & \textbf{0.798} & \textbf{0.9169} & \textbf{0.3263} & \textbf{0.4477} & \textbf{0.8016} & 0.1819 & 0.3508 & 0.8767 & 0.3339 & 0.4609 & 0.7938 \\
& GAT & 0.4762 & 1.1623 & 0.839 & 1.0883 & 1.8336 & 0.5611 & 0.4834 & 0.6394 & 0.5958 & 0.1812 & 0.3374 & 0.8861 & 0.6848 & 0.8683 & 0.2541 \\
\midrule
\multirow{3}{*}{Validation} 
& GINE & \textbf{0.6332} & \textbf{1.2972} & \textbf{0.8219} & 1.095 & 1.6276 & 0.6929 & \textbf{0.3604} & \textbf{0.4516} & \textbf{0.748} & 0.4134 & 0.7032 & 0.5137 & 0.5431 & 0.6576 & 0.4674 \\
& ResGatedGCN & 0.7485 & 1.4158 & 0.7879 & \textbf{0.8544} & \textbf{1.2884} & \textbf{0.8075} & 0.3772 & 0.4809 & 0.7142 & \textbf{0.3637} & \textbf{0.5815} & \textbf{0.6675} & \textbf{0.4462} & \textbf{0.5689} & \textbf{0.6108} \\
& GAT & 1.1039 & 2.2889 & 0.4455 & 1.9788 & 2.3872 & 0.3393 & 0.4518 & 0.5596 & 0.6131 & 0.4071 & 0.7779 & 0.4049 & 0.7117 & 0.8487 & 0.2643 \\
\midrule
\multirow{3}{*}{Test} 
& GINE & \textbf{0.3468} & \textbf{1.0331} & \textbf{0.8878} & 0.7194 & 1.3095 & 0.8003 & \textbf{0.4337} & \textbf{0.5804} & \textbf{0.7109} & 0.4755 & 0.9655 & -0.0107 & 0.4903 & 0.6615 & 0.6151 \\
& ResGatedGCN & 0.3694 & 1.0668 & 0.8803 & \textbf{0.6355} & \textbf{1.1277} & \textbf{0.8519} & 0.4576 & 0.6043 & 0.6866 & 0.4930 & 0.9892 & -0.0610 & \textbf{0.4717} & \textbf{0.6207} & \textbf{0.7067} \\
& GAT & 0.5068 & 1.3621 & 0.8049 & 1.6859 & 2.6390 & 0.1889 & 0.6109 & 0.7568 & 0.5085 & \textbf{0.4276} & \textbf{0.9374} & \textbf{0.0472} & 0.8857 & 1.0320 & 0.0634 \\
\midrule
\multicolumn{2}{c|}{\textbf{Avg.\ Test}} & \textbf{0.4077} & \textbf{1.1540} & \textbf{0.8577} & 1.0136 & 1.6921 & 0.6137 & 0.5007 & 0.6472 & 0.6353 & 0.4654 & 0.9640 & $-$0.0082 & 0.6159 & 0.7714 & 0.4617 \\
\bottomrule
\end{tabular}}
\end{table*}

Metrics follow MAE/RMSE/R\textsuperscript{2}; lower MAE/RMSE and higher R\textsuperscript{2} indicate better performance. Train/Validation/Test splits are merged into a single wide table for side-by-side comparison (\autoref{tab:placement_views}). The Avg.\ Test R\textsuperscript{2} row reports the mean across the three GNN baselines, providing a view-level summary. Intuitively, node-centric graphs align more directly with gate-position labels than edge-only formulations or pin-incidence encodings, helping explain the stronger generalization of view~(b).
Critically, node-centric views~(b/c) are preferred by all classic GNNs, consistently surpassing edge-only views~(d/e) and the widely used AIG-equivalent view~(f).

Across views, view~(b) is consistently strongest on Test R\textsuperscript{2}, with view~(c) close for ResGatedGCN. View~(d) is mid-tier; view~(e) collapses for GINE/ResGatedGCN; and view~(f) is moderate. These trends are mirrored by MAE/RMSE, indicating practical significance beyond variance.
Importantly, view~(b) fits the physical design stages well and outperforms view~(f) by large margins, suggesting view~(f) should not be the default representation for learning-based placement.

Across models, performance is view-dependent: on view~(b), GINE slightly edges ResGatedGCN; on view~(c), ResGatedGCN surpasses GINE; on view~(d), GINE $>$ ResGatedGCN; on view~(e), both GINE and ResGatedGCN are negative whereas GAT is barely positive; on view~(f), ResGatedGCN $>$ GINE $\gg$ GAT. The pattern suggests that residual gating stabilizes edge-aware message passing on node-centric graphs, while attention-only GAT underperforms in geometry-coupled regimes.
Our \textit{critical finding} lies here: for a fixed GNN, view choice affects performance far more than expected; rankings turn over across views. For example, GAT achieves its best accuracy only on view~(e).

Negative R\textsuperscript{2} on view~(e) indicates a label--view mismatch: incidence-edge supervision dilutes spatial inductive bias needed for node targets. Moderate performance on view~(f) suggests it can serve as auxiliary supervision rather than a primary view. This pattern underscores the importance of matching representation to target geometry.

With placement trends established, we next examine whether these view-dependent patterns persist when labels are connectivity-driven.

\subsection{Comparison of Different Views for Routing Tasks}

\begin{table*}[t]
\centering
\caption{Routing results across views (b--f) and splits. Lower MAE/RMSE is better; higher R\textsuperscript{2} is better.}
\label{tab:routing_results}
\resizebox{\linewidth}{!}{%
\begin{tabular}{ll| 
    >{\columncolor[HTML]{E0E0E0}}c >{\columncolor[HTML]{E0E0E0}}c >{\columncolor[HTML]{E0E0E0}}c 
    >{\columncolor[HTML]{F9F2E7}}c >{\columncolor[HTML]{F9F2E7}}c >{\columncolor[HTML]{F9F2E7}}c 
    >{\columncolor[HTML]{E6F2FF}}c >{\columncolor[HTML]{E6F2FF}}c >{\columncolor[HTML]{E6F2FF}}c 
    >{\columncolor[HTML]{E6FFE6}}c >{\columncolor[HTML]{E6FFE6}}c >{\columncolor[HTML]{E6FFE6}}c 
    >{\columncolor[HTML]{FFF7E6}}c >{\columncolor[HTML]{FFF7E6}}c >{\columncolor[HTML]{FFF7E6}}c 
    }
\toprule
& & \multicolumn{3}{c}{\cellcolor[HTML]{E0E0E0}\textbf{(b)}} & \multicolumn{3}{c}{\cellcolor[HTML]{F9F2E7}\textbf{(c)}} & \multicolumn{3}{c}{\cellcolor[HTML]{E6F2FF}\textbf{(d)}} & \multicolumn{3}{c}{\cellcolor[HTML]{E6FFE6}\textbf{(e)}} & \multicolumn{3}{c}{\cellcolor[HTML]{FFF7E6}\textbf{(f)}} \\ 
\cmidrule(lr){3-5} \cmidrule(lr){6-8} \cmidrule(lr){9-11} \cmidrule(lr){12-14} \cmidrule(lr){15-17}
\textbf{Split} & \textbf{Model} & \textbf{MAE $\downarrow$} & \textbf{RMSE $\downarrow$} & \textbf{R\textsuperscript{2} $\uparrow$} & \textbf{MAE $\downarrow$} & \textbf{RMSE $\downarrow$} & \textbf{R\textsuperscript{2} $\uparrow$} & \textbf{MAE $\downarrow$} & \textbf{RMSE $\downarrow$} & \textbf{R\textsuperscript{2} $\uparrow$} & \textbf{MAE $\downarrow$} & \textbf{RMSE $\downarrow$} & \textbf{R\textsuperscript{2} $\uparrow$} & \textbf{MAE $\downarrow$} & \textbf{RMSE $\downarrow$} & \textbf{R\textsuperscript{2} $\uparrow$}\\
\midrule
\multirow{3}{*}{Train}
& GINE & \textbf{0.2520} & 0.7796 & 0.9257 & 0.7011 & 1.3372 & 0.8716 & 0.4019 & 0.5419 & 0.7105 & 0.4188 & 0.5744 & 0.6746 & \textbf{0.1477} & \textbf{0.2130} & \textbf{0.9551} \\
& ResGatedGCN & 0.3041 & \textbf{0.6848} & \textbf{0.9427} & \textbf{0.5291} & \textbf{0.9252} & \textbf{0.9385} & \textbf{0.3947} & \textbf{0.5400} & \textbf{0.7126} & 0.3805 & 0.5217 & 0.7318 & 0.5545 & 0.7204 & 0.4886 \\
& GAT & 0.3944 & 1.0904 & 0.8547 & 1.4430 & 2.5930 & 0.5170 & 0.5659 & 0.7216 & 0.4871 & \textbf{0.2914} & \textbf{0.4246} & \textbf{0.8224} & 0.8544 & 1.0354 & -0.0618 \\
\midrule
\multirow{3}{*}{Validation}
& GINE & 0.8523 & 1.7891 & 0.6529 & 1.7403 & 2.8566 & 0.4877 & \textbf{0.3892} & \textbf{0.5013} & \textbf{0.6601} & 0.3839 & 0.5155 & \textbf{0.6370} & \textbf{0.2670} & \textbf{0.4098} & \textbf{0.8143} \\
& ResGatedGCN & \textbf{0.7971} & \textbf{1.5267} & \textbf{0.7472} & \textbf{1.0107} & \textbf{1.7078} & \textbf{0.8169} & 0.4222 & 0.5288 & 0.6218 & 0.3929 & 0.5237 & 0.6255 & 0.6243 & 0.7871 & 0.1746 \\
& GAT & 1.1039 & 2.2035 & 0.4734 & 2.6724 & 3.3680 & 0.2879 & 0.5718 & 0.7148 & 0.3089 & \textbf{0.1054} & \textbf{0.3233} & 0.4816 & 0.8053 & 0.9564 & -0.0111 \\
\midrule
\multirow{3}{*}{Test}
& GINE & \textbf{0.2994} & 1.1373 & 0.8596 & 1.2648 & 2.3667 & 0.6471 & \textbf{0.5746} & \textbf{0.7391} & \textbf{0.5758} & 0.6993 & 0.8588 & 0.4277 & 0.8106 & 0.9531 & 0.2972 \\
& ResGatedGCN & 0.3498 & \textbf{1.0676} & \textbf{0.8763} & \textbf{0.7856} & \textbf{1.4808} & \textbf{0.8618} & 0.5912 & 0.7640 & 0.5467 & 0.5001 & 0.6548 & 0.6673 & \textbf{0.7634} & \textbf{0.9402} & \textbf{0.3179} \\
& GAT & 0.4477 & 1.3123 & 0.8131 & 2.2493 & 3.7245 & 0.1260 & 0.8769 & 1.1125 & 0.0389 & \textbf{0.3202} & \textbf{0.4557} & \textbf{0.8389} & 1.0043 & 1.1361 & 0.0014 \\
\midrule
\multicolumn{2}{c|}{\textbf{Avg.\ Test}} & \textbf{0.3656} & \textbf{1.1724} & \textbf{0.8497} & 1.4332 & 2.5240 & 0.5450 & 0.6809 & 0.8719 & 0.3871 & 0.5065 & 0.6564 & 0.6446 & 0.8594 & 1.0098 & 0.2055 \\
\bottomrule
\end{tabular}}
\end{table*}

Routing is connectivity-driven; consequently, incidence and net-edge formulations (views~(e) and (d)) should be competitive, while view~(b) remains strong due to coupled geometry and features. Using the same models and metrics, \autoref{tab:routing_results} reports Train/Validation/Test performance across views~(b--f).

Despite connectivity-centric labels, the node-centric view~(b) remains the strongest across baselines, reinforcing the general preference for node-centric representations.

Across views, view~(b) shows the best generalization on the test split. Among edge-based views, view~(e) performs strongest overall, with view~(d) remaining competitive, while view~(f) consistently underperforms. These trends are consistent across data splits.

Across models, architectures exhibit clear view preferences. ResGatedGCN performs best on node-centric views (b,c), GINE remains competitive across views (b--d), and GAT performs best on view~(e) but degrades on other views.

Our \textit{critical finding} is that view choice dominates model choice: model rankings change substantially across graph representations. Attention-based models perform well on view~(e) but lag on views~(b--d), while GINE and ResGatedGCN alternate as the strongest models depending on the representation.

The strong performance of GAT on view~(e) suggests that attention benefits from incidence-edge neighborhoods capturing dense local connectivity, whereas pairwise nets in view~(d) create sparser structures that weaken attention aggregation. ResGatedGCN remains stable on views~(b,c), likely due to residual-gated aggregation that preserves geometric coupling. Generalization gaps widen on edge-only views, indicating weaker alignment between connectivity-only inputs and routing labels.

Compared with placement, routing is generally more challenging for classical GNNs, with lower predictive accuracy and larger generalization gaps.

Taken together, view~(b) is the most reliable representation for routing tasks, while view~(e) serves as the strongest edge-oriented complement. View~(f) should generally be avoided. Model selection should match the representation: ResGatedGCN is most reliable on views~(b,c), whereas GAT can be effective on view~(e) but is less stable across views.
\subsection{Comparison of GNNs and Task-Specific Heads}

We now turn from view-level recommendations to model capacity: how message-passing depth and decoder-head design drive performance.

Building on these findings, on node-centric view~(b), a moderate message-passing depth (3--4 layers) suffices; deeper stacks risk over-smoothing and optimization decay \cite{li2018deeper}. More importantly, \emph{decoder-head configuration} is the dominant lever for placement and routing accuracy, a dimension underexplored in graph ML/EDA benchmarks that primarily tune the message-passing stack \cite{hu2020open,Dwivedi2020BenchmarkingGNNs,xu2018powerful,bresson2017residual}. With information parity across multi-view, typed circuit graphs and stage- and resolution-matched targets, R2G reveals that increasing head depth from 1 to 3--4 layers transforms performance and training stability (often flipping model rankings). With this context, we first summarize the \emph{layer} ablations below, then return to \emph{head} effects.
\subsubsection{GNN Layers vs. Accuracy}

\begin{table}[tbp]
\centering
\caption{Placement on view~(b) vs. GNN depth (3--6). Compact stacks (3--4) yield highest accuracy; deeper (6) underperform.}
\label{tab:place_gnnlayers}
\resizebox{\linewidth}{!}{%
\begin{tabular}{c| 
    >{\columncolor[HTML]{FFF7E6}}c >{\columncolor[HTML]{FFF7E6}}c >{\columncolor[HTML]{F9F2E7}}c >{\columncolor[HTML]{FFF7E6}}c 
    >{\columncolor[HTML]{E6F2FF}}c >{\columncolor[HTML]{E6F2FF}}c >{\columncolor[HTML]{E6F2FF}}c >{\columncolor[HTML]{E6F2FF}}c 
    }
\toprule
& \multicolumn{4}{c}{\cellcolor[HTML]{FFF7E6}\textbf{GINE}} & \multicolumn{4}{c}{\cellcolor[HTML]{E6F2FF}\textbf{ResGatedGCN}} \\
\cmidrule(lr){2-5} \cmidrule(lr){6-9}
\textbf{\#layers} & \textbf{MAE$\downarrow$} & \textbf{RMSE$\downarrow$} & \textbf{R²$\uparrow$} & \textbf{\#Param.} & \textbf{MAE$\downarrow$} & \textbf{RMSE$\downarrow$} & \textbf{R²$\uparrow$} & \textbf{\#Param.} \\
\midrule
3 & 0.3489 & 1.0454 & 0.8851 & 0.78 M & 0.3319 & \textbf{1.0296} & \textbf{0.8885} & 0.98 M \\
4 & \textbf{0.2594} & \textbf{1.0158} & \textbf{0.8915} & 0.98 M & 0.3249 & 1.0617 & 0.8815 & 1.24 M \\
5 & 0.2991 & 1.0445 & 0.8853 & 1.18 M & \textbf{0.2923} & 1.0699 & 0.8797 & 1.51 M \\
6 & 0.3312 & 1.2536 & 0.8348 & 1.38 M & 0.3327 & 1.0833 & 0.8766 & 1.77 M \\
\bottomrule
\end{tabular}
}
\end{table}

\begin{table}[tbp]
\centering
\caption{Routing on view~(b) vs. GNN depth (3--6). Performance peaks at 3 layers for both architectures; deeper stacks show diminishing returns.}
\label{tab:route_gnnlayers_test}
\resizebox{\linewidth}{!}{%
\begin{tabular}{c| 
    >{\columncolor[HTML]{FFF7E6}}c >{\columncolor[HTML]{FFF7E6}}c >{\columncolor[HTML]{FFF7E6}}c >{\columncolor[HTML]{FFF7E6}}c 
    >{\columncolor[HTML]{E6F2FF}}c >{\columncolor[HTML]{E6F2FF}}c >{\columncolor[HTML]{E6F2FF}}c >{\columncolor[HTML]{E6F2FF}}c 
    }
\toprule
& \multicolumn{4}{c}{\cellcolor[HTML]{FFF7E6}\textbf{GINE}} & \multicolumn{4}{c}{\cellcolor[HTML]{E6F2FF}\textbf{ResGatedGCN}} \\
\cmidrule(lr){2-5} \cmidrule(lr){6-9}
\textbf{\#layers} & \textbf{MAE$\downarrow$} & \textbf{RMSE$\downarrow$} & \textbf{R²$\uparrow$} & \textbf{\#Param.} & \textbf{MAE$\downarrow$} & \textbf{RMSE$\downarrow$} & \textbf{R²$\uparrow$} & \textbf{\#Param.} \\
\midrule
3 & 0.3220 & \textbf{1.0691} & \textbf{0.8759} & 0.78 M & 0.3439 & \textbf{1.0682} & \textbf{0.8762} & 0.98 M \\
4 & 0.3466 & 1.1503 & 0.8664 & 0.98 M & \textbf{0.2977} & 1.1141 & 0.8653 & 1.24 M \\
5 & \textbf{0.2845} & 1.0969 & 0.8694 & 1.18 M & 0.3188 & 1.0768 & 0.8742 & 1.51 M \\
6 & 0.3670 & 1.0910 & 0.8708 & 1.38 M & 0.3892 & 1.1495 & 0.8566 & 1.77 M \\
\bottomrule
\end{tabular}
}
\end{table}

For placement on view~(b), accuracy varies non-monotonically with depth (\autoref{tab:place_gnnlayers}). Both GINE and ResGatedGCN achieve their best performance at moderate depths and decline as depth increases, indicating diminishing returns.

Moderate depth balances receptive-field coverage and optimization stability, whereas deeper stacks risk over-smoothing and gradient decay. The sharper degradation observed for GINE suggests greater depth sensitivity of its aggregation mechanism. Consequently, compact stacks (3--4 layers) are preferred for node-centric placement on view~(b).

Turning to routing (\autoref{tab:route_gnnlayers_test}), we observe a similar non-monotonic pattern: both GINE and ResGatedGCN perform best at moderate depths and degrade as depth increases. Excessive depth diffuses signal across nets and increases training variance; the drop observed for ResGatedGCN at intermediate depths may indicate gating saturation. Accordingly, compact stacks (3--4 layers) are also preferred for routing on view~(b).
\subsubsection{Head Layers vs. Accuracy}

\begin{table}[tb]
  \centering
  \caption{Placement on view~(b) vs. head layers (1--4). Increasing head capacity stabilizes training and drives accuracy; 3--4 head layers deliver near-perfect R$^2$ across architectures.}
  \label{tab:place_headlayers_test}
  \resizebox{\linewidth}{!}{%
\begin{tabular}{c| 
    >{\columncolor[HTML]{FFF7E6}}c >{\columncolor[HTML]{FFF7E6}}c >{\columncolor[HTML]{FFF7E6}}c >{\columncolor[HTML]{FFF7E6}}c 
    >{\columncolor[HTML]{E6F2FF}}c >{\columncolor[HTML]{E6F2FF}}c >{\columncolor[HTML]{E6F2FF}}c >{\columncolor[HTML]{E6F2FF}}c 
    }
\toprule
& \multicolumn{4}{c}{\cellcolor[HTML]{FFF7E6}\textbf{GINE}} & \multicolumn{4}{c}{\cellcolor[HTML]{E6F2FF}\textbf{ResGatedGCN}} \\
\cmidrule(lr){2-5} \cmidrule(lr){6-9}
\textbf{\#layers} & MAE$\downarrow$ & RMSE$\downarrow$ & R\textsuperscript{2}$\uparrow$ & \#Param. & MAE$\downarrow$ & RMSE$\downarrow$ & R\textsuperscript{2}$\uparrow$ & \#Param. \\
\midrule
1 & 0.4708 & 1.1886 & 0.8514 & 0.91 M & 3.1580 & 3.3342 & -0.1689 & 1.18 M \\
2 & 0.3414 & 1.1184 & 0.8685 & 0.98 M & 0.3249 & 1.0542 & 0.8832 & 1.24 M \\
3 & 0.0870 & 0.3329 & 0.9883 & 1.05 M & \textbf{0.0963} & 0.3251 & 0.9889 & 1.31 M \\
4 & \textbf{0.0703} & \textbf{0.1909} & \textbf{0.9962} & 1.11 M & 0.1421 & \textbf{0.3223} & \textbf{0.9891} & 1.38 M \\
\bottomrule
  \end{tabular}
  }
\end{table}

\begin{table}[tb]
  \centering
  \caption{Routing on view~(b) vs. head layers (1--4). Shallow heads are unstable; accuracy improves sharply at 3 and is near-perfect at 4. Prefer 3--4 head layers for robust routing.}
  \label{tab:route_headlayers_test}
  \resizebox{\linewidth}{!}{%
\begin{tabular}{c| 
    >{\columncolor[HTML]{FFF7E6}}c >{\columncolor[HTML]{FFF7E6}}c >{\columncolor[HTML]{FFF7E6}}c >{\columncolor[HTML]{FFF7E6}}c 
    >{\columncolor[HTML]{E6F2FF}}c >{\columncolor[HTML]{E6F2FF}}c >{\columncolor[HTML]{E6F2FF}}c >{\columncolor[HTML]{E6F2FF}}c 
    }
\toprule
& \multicolumn{4}{c}{\cellcolor[HTML]{FFF7E6}\textbf{GINE}} & \multicolumn{4}{c}{\cellcolor[HTML]{E6F2FF}\textbf{ResGatedGCN}} \\
\cmidrule(lr){2-5} \cmidrule(lr){6-9}
\textbf{\#layers} & MAE$\downarrow$ & RMSE$\downarrow$ & R\textsuperscript{2}$\uparrow$ & \#Param. & MAE$\downarrow$ & RMSE$\downarrow$ & R\textsuperscript{2}$\uparrow$ & \#Param. \\
\midrule
1 & 13.9757 & 54.1022 & nan & 0.91 M & nan & nan & nan & 1.18 M \\
2 & 0.3425 & 1.1414 & 0.8586 & 0.98 M & 0.2892 & 1.1276 & 0.8620 & 1.24 M \\
3 & 0.0761 & 0.3173 & 0.9891 & 1.05 M & \textbf{0.1230} & \textbf{0.3823} & \textbf{0.9841} & 1.31 M \\
4 & \textbf{0.0542} & \textbf{0.1802} & \textbf{0.9965} & 1.11 M & 0.2482 & 0.3824 & \textbf{0.9841} & 1.38 M \\
\bottomrule
  \end{tabular}
  }
\end{table}

On placement, performance rises sharply with head depth: moving from head 1 to head 3 lifts Test R\textsuperscript{2} from 0.8514 to 0.9883 (GINE) and from $-0.1689$ to 0.9889 (ResGatedGCN); head 4 reaches near-perfect accuracy for GINE (0.9962) and maintains or slightly improves ResGatedGCN (0.9891).
Moreover, shallow heads cannot linearize the mapping from node features to coordinates; additional depth provides the nonlinearity and normalization capacity needed for stable generalization. The negative R\textsuperscript{2} at head 1 for ResGatedGCN is a clear outlier and reflects under-parameterization and poor calibration.
Consequently, use 3--4 head layers on view~(b); this yields stable, near-perfect placement accuracy.

For routing, head 1 is unstable (\texttt{nan} metrics); accuracy improves sharply at head 3 and reaches near-perfect at head 4 (GINE 0.9965; ResGatedGCN 0.9841).
Furthermore, insufficient head capacity causes optimization failure and numerical instability; nan metrics at head 1 confirm this. Routing labels encode local connectivity patterns requiring nonlinear fusion, which shallow heads cannot provide.
Accordingly, adopt 3--4 head layers for routing on view~(b); shallow heads are brittle and should be avoided.

Overall, on node-centric view~(b), compact message-passing depth (3--4 layers) suffices, while deeper stacks show over-smoothing and diminishing returns \cite{li2018deeper}. In contrast, \emph{decoder-head capacity} is the primary driver of accuracy and stability: increasing the head from shallow settings to 3--4 layers consistently turns unstable optimization, particularly for routing, into accurate predictions. This pattern holds across architectures and can flip model rankings depending on view and head choices. Supplementary experiments on view~(d) show the same qualitative behavior for edge-centric settings, as reported in \autoref{sec:gnn_head}.

R2G’s uniqueness lies in its controlled, multi-view, stage- and resolution-matched design with information parity, which decouples representation from modeling and makes decoder-head effects visible in ways that prior benchmarks, which emphasize message-passing stacks, seldom do \cite{hu2020open,xu2018powerful}. Practically, pair view~(b) with 3--4 GNN layers and 3--4 head layers for robust, high accuracy, and prioritize view/head configuration before increasing depth.

\section{Conclusions \& Future Work}
We have presented R2G, the first multi-view circuit-graph benchmark suite for physical design, providing five stage-aware views with information parity over 30 open-source IP cores, together with an end-to-end DEF-to-graph pipeline, loaders, unified splits, domain metrics, and reproducible baselines. Experiments with classic GNNs yield three transferable findings: (i)~view choice dominates model choice, with Test R$^2$ varying by more than 0.3 across representations for a fixed GNN; (ii)~node-centric views generalize best across both placement and routing, with effective views aligning with supervision granularity; and (iii)~decoder-head depth (3--4 layers) is the primary accuracy driver, turning divergent training into near-perfect predictions (R$^2$$>$0.99), a finding prior benchmarks have not surfaced. For practitioners, we recommend view~(b) as the default representation, 3--4 GNN layers to avoid over-smoothing, and prioritizing decoder-head configuration before increasing model depth.

\textbf{Limitations and future work.} R2G currently covers late physical-design stages; DEF-derived attributes omit timing, IR-drop, and detailed power; designs span limited technology nodes; and evaluations center on classic GNNs. Future work includes broader design and technology coverage, timing- and congestion-aware tasks with richer LEF/STA/PDN attributes, graph transformers and MoE architectures, and stronger EDA evaluation protocols for broader adoption.

\bibliographystyle{ieeenat_fullname}
\bibliography{references}

@article{Pan2023EDALearnAC,
  title={EDALearn: A Comprehensive RTL-to-Signoff EDA Benchmark for Democratized and Reproducible ML for EDA Research},
  author={Jingyu Pan and Chen-Chia Chang and Zhiyao Xie and Yiran Chen and Hai Helen Li},
  journal={Proceedings of the 43rd IEEE/ACM International Conference on Computer-Aided Design},
  year={2023},
  url={https://api.semanticscholar.org/CorpusID:265609553}
}

@misc{openroad_docs,
  title        = {OpenROAD Documentation},
  howpublished = {\url{https://openroad.readthedocs.io/en/latest/main/README.html}},
  note         = {Accessed: 2025-11-06},
}

@misc{openroad_github,
  title        = {OpenROAD: Unified Application for RTL-to-GDSII Flow},
  howpublished = {\url{https://github.com/The-OpenROAD-Project/OpenROAD}},
  note         = {Accessed: 2025-11-06},
}

@article{chai2022circuitnet,
  title={Circuitnet: An open-source dataset for machine learning applications in electronic design automation (eda)},
  author={Chai, Zhuomin and Zhao, Yuxiang and Lin, Yibo and Liu, Wei and Wang, Runsheng and Huang, Ru},
  journal={arXiv preprint arXiv:2208.01040},
  year={2022}
}

@article{shi2025forgeeda,
  title={ForgeEDA: A Comprehensive Multimodal Dataset for Advancing EDA},
  author={Shi, Zhengyuan and Li, Zeju and Ma, Chengyu and Zhou, Yunhao and Zheng, Ziyang and Liu, Jiawei and Pan, Hongyang and Zhou, Lingfeng and Li, Kezhi and Zhu, Jiaying and others},
  journal={arXiv preprint arXiv:2505.02016},
  year={2025}
}

@inproceedings{shrestha2024edaschema,
  title={EDA-schema: A Graph Datamodel Schema and Open Dataset for Digital Design Automation},
  author={Shrestha, Pratik and Aversa, Alec and Phatharodom, Saran and Savidis, Ioannis},
  booktitle={Proceedings of the Great Lakes Symposium on VLSI 2024},
  pages={69--77},
  year={2024},
  doi={10.1145/3649476.3658718}
}

@inproceedings{guo2022timingpredict,
  title={A timing engine inspired graph neural network model for pre-routing slack prediction},
  author={Guo, Zizheng and Liu, Mingjie and Gu, Jiaqi and Zhang, Shuhan and Pan, David Z and Lin, Yibo},
  booktitle={Proceedings of the 59th ACM/IEEE Design Automation Conference},
  pages={1207--1212},
  year={2022},
  doi={10.1145/3489517.3530597}
}

@inproceedings{luo2024dehnn,
  title={{DE-HNN}: An effective neural model for Circuit Netlist representation},
  author={Luo, Zhishang and Son Hy, Truong and Tabaghi, Puoya and Defferrard, Micha{\"e}l and Rezaei, Elahe and Carey, Ryan M and Davis, Rhett and Jain, Rajeev and Wang, Yusu},
  booktitle={Proceedings of The 27th International Conference on Artificial Intelligence and Statistics},
  pages={4258--4266},
  year={2024},
  volume={238},
  series={Proceedings of Machine Learning Research},
  url={https://proceedings.mlr.press/v238/luo24a.html}
}

@article{ni2024openlsdgf,
  title={OpenLS-DGF: An Adaptive Open-Source Dataset Generation Framework for Machine Learning Tasks in Logic Synthesis},
  author={Ni, Liwei and Wang, Rui and Liu, Miao and Meng, Xingyu and Lin, Xiaoze and Liu, Junfeng and Luo, Guojie and Chu, Zhufei and Qian, Weikang and Yang, Xiaoyan and Xie, Biwei and Li, Xingquan and Li, Huawei},
  journal={arXiv preprint arXiv:2411.09422},
  year={2024},
  url={https://arxiv.org/abs/2411.09422}
}

@article{alrahis2022gnnre,
  title={{GNN-RE}: Graph Neural Networks for Reverse Engineering of Gate-Level Netlists},
  author={Alrahis, Lilas and Sengupta, Abhrajit and Knechtel, Johann and Patnaik, Satwik and Saleh, Hani and Mohammad, Baker and Al-Qutayri, Mahmoud and Sinanoglu, Ozgur},
  journal={IEEE Transactions on Computer-Aided Design of Integrated Circuits and Systems},
  volume={41},
  number={8},
  pages={2435--2448},
  year={2022},
  publisher={IEEE},
  doi={10.1109/TCAD.2021.3110807}
}

@inproceedings{dong2023cktgnn,
  title={CktGNN: Circuit Graph Neural Network for Electronic Design Automation},
  author={Dong, Zehao and Cao, Weidong and Zhang, Muhan and Tao, Dacheng and Chen, Yixin and Zhang, Xuan},
  booktitle={The Eleventh International Conference on Learning Representations},
  year={2023},
  url={https://openreview.net/forum?id=NE2911Kq1sp}
}

@inproceedings{tao2024amsnet,
  title={AMSNet: Netlist Dataset for AMS Circuits},
  author={Tao, Zhuofu and Shi, Yichen and Huo, Yiru and Ye, Rui and Li, Zonghang and Huang, Li and Wu, Chen and Bai, Na and Yu, Zhiping and Lin, Ting-Jung and He, Lei},
  booktitle={2024 IEEE LLM Aided Design Workshop (LAD)},
  pages={1--5},
  year={2024},
  organization={IEEE},
  doi={10.1109/LAD62341.2024.10691781}
}

@article{hutton2002characterization,
  title={Characterization and parameterized generation of synthetic combinational benchmark circuits},
  author={Hutton, Michael D and Rose, Jonathan and Grossman, Jerry P and Corneil, Derek G},
  journal={IEEE Transactions on Computer-Aided Design of Integrated Circuits and Systems},
  volume={17},
  number={10},
  pages={985--996},
  year={2002},
  publisher={IEEE}
}

@article{chowdhury2021openabc,
  title={Openabc-d: A large-scale dataset for machine learning guided integrated circuit synthesis},
  author={Chowdhury, Animesh Basak and Tan, Benjamin and Karri, Ramesh and Garg, Siddharth},
  journal={arXiv preprint arXiv:2110.11292},
  year={2021}
}

@inproceedings{jiang2024circuitnet,
  title={Circuitnet 2.0: An advanced dataset for promoting machine learning innovations in realistic chip design environment},
  author={Jiang, Xun and Zhao, Yuxiang and Lin, Yibo and Wang, Runsheng and Huang, Ru and others},
  booktitle={The Twelfth International Conference on Learning Representations},
  year={2024}
}

@article{liao2023dreamplace,
  title={Dreamplace 4.0: Timing-driven placement with momentum-based net weighting and lagrangian-based refinement},
  author={Liao, Peiyu and Guo, Dawei and Guo, Zizheng and Liu, Siting and Lin, Yibo and Yu, Bei},
  journal={IEEE Transactions on Computer-Aided Design of Integrated Circuits and Systems},
  volume={42},
  number={10},
  pages={3374--3387},
  year={2023},
  publisher={IEEE}
}

@inproceedings{shen2024deep,
  title={Deep-Learning-Based Pre-Layout Parasitic Capacitance Prediction on SRAM Designs},
  author={Shen, Shan and Yang, Dingcheng and Xie, Yuyang and Pei, Chunyan and Yu, Bei and Yu, Wenjian},
  booktitle={Proceedings of the Great Lakes Symposium on VLSI 2024},
  pages={440--445},
  year={2024}
}

@article{yang2023cnn,
  title={CNN-Cap: Effective convolutional neural network-based capacitance models for interconnect capacitance extraction},
  author={Yang, Dingcheng and Li, Haoyuan and Yu, Wenjian and Guo, Yuanbo and Liang, Wenjie},
  journal={ACM Transactions on Design Automation of Electronic Systems},
  volume={28},
  number={4},
  pages={1--22},
  year={2023},
  publisher={ACM New York, NY}
}

@inproceedings{jiang2023accelerating,
  title={Accelerating routability and timing optimization with open-source ai4eda dataset circuitnet and heterogeneous platforms},
  author={Jiang, Xun and Guo, Zizheng and Chai, Zhuomin and Zhao, Yuxiang and Lin, Yibo and Wang, Runsheng and Huang, Ru},
  booktitle={2023 IEEE/ACM International Conference on Computer Aided Design (ICCAD)},
  pages={1--9},
  year={2023},
  organization={IEEE}
}

@article{hu2020open,
  title={Open graph benchmark: Datasets for machine learning on graphs},
  author={Hu, Weihua and Fey, Matthias and Zitnik, Marinka and Dong, Yuxiao and Ren, Hongyu and Liu, Bowen and Catasta, Michele and Leskovec, Jure},
  journal={Advances in neural information processing systems},
  volume={33},
  pages={22118--22133},
  year={2020}
}

@article{xu2018powerful,
  title={How powerful are graph neural networks?},
  author={Xu, Keyulu and Hu, Weihua and Leskovec, Jure and Jegelka, Stefanie},
  journal={arXiv preprint arXiv:1810.00826},
  year={2018}
}

@article{bresson2017residual,
  title={Residual gated graph convnets},
  author={Bresson, Xavier and Laurent, Thomas},
  journal={arXiv preprint arXiv:1711.07553},
  year={2017}
}

@inproceedings{li2018deeper,
  title={Deeper insights into graph convolutional networks for semi-supervised learning},
  author={Li, Qimai and Han, Zhichao and Wu, Xiao-Ming},
  booktitle={Thirty-Second AAAI conference on artificial intelligence},
  year={2018}
}

@article{morris2020tudataset,
  title={Tudataset: A collection of benchmark datasets for learning with graphs},
  author={Morris, Christopher and Kriege, Nils M and Bause, Franka and Kersting, Kristian and Mutzel, Petra and Neumann, Marion},
  journal={arXiv preprint arXiv:2007.08663},
  year={2020}
}

@article{Dwivedi2020BenchmarkingGNNs,
  title={Benchmarking Graph Neural Networks},
  author={Dwivedi, Vijay Prakash and Joshi, Chaitanya K and Laurent, Thomas and Bengio, Yoshua and Bresson, Xavier},
  journal={arXiv preprint arXiv:2003.00982},
  year={2020}
}

@article{Sterling2015ZINC15,
  title={ZINC 15—Ligand Discovery for Everyone},
  author={Sterling, Teague and Irwin, John J.},
  journal={Journal of Chemical Information and Modeling},
  volume={55},
  number={11},
  pages={2324--2337},
  year={2015}
}

@article{Ramakrishnan2014QM9,
  title={Quantum chemistry structures and properties of 134k molecules},
  author={Ramakrishnan, Raghunathan and Dral, Pavlo O. and Rupp, Matthias and von Lilienfeld, O. Anatole},
  journal={Scientific Data},
  volume={1},
  pages={140022},
  year={2014}
}

@inproceedings{shen2025few,
  title={Few-shot learning on ams circuits and its application to parasitic capacitance prediction},
  author={Shen, Shan and Zhang, Yibin and Rodriguez, Hector Rodriguez and Yu, Wenjian},
  booktitle={2025 62nd ACM/IEEE Design Automation Conference (DAC)},
  pages={1--7},
  year={2025},
  organization={IEEE}
}

@article{yang2022versatile,
  title={Versatile multi-stage graph neural network for circuit representation},
  author={Yang, Shuwen and Yang, Zhihao and Li, Dong and Zhang, Yingxueff and Zhang, Zhanguang and Song, Guojie and Hao, Jianye},
  journal={Advances in Neural Information Processing Systems},
  volume={35},
  pages={20313--20324},
  year={2022}
}

@article{gao2025analoggenie,
  title={AnalogGenie: A generative engine for automatic discovery of analog circuit topologies},
  author={Gao, Jian and Cao, Weidong and Yang, Junyi and Zhang, Xuan},
  journal={arXiv preprint arXiv:2503.00205},
  year={2025}
}

@inproceedings{shi2023deepgate2,
  title={Deepgate2: Functionality-aware circuit representation learning},
  author={Shi, Zhengyuan and Pan, Hongyang and Khan, Sadaf and Li, Min and Liu, Yi and Huang, Junhua and Zhen, Hui-Ling and Yuan, Mingxuan and Chu, Zhufei and Xu, Qiang},
  booktitle={2023 IEEE/ACM International Conference on Computer Aided Design (ICCAD)},
  pages={1--9},
  year={2023},
  organization={IEEE}
}

@inproceedings{shi2024deepgate3,
  title={Deepgate3: Towards scalable circuit representation learning},
  author={Shi, Zhengyuan and Zheng, Ziyang and Khan, Sadaf and Zhong, Jianyuan and Li, Min and Xu, Qiang},
  booktitle={Proceedings of the 43rd IEEE/ACM International Conference on Computer-Aided Design},
  pages={1--9},
  year={2024}
}

@misc{cora_website,
  title        = {The Cora dataset},
  author       = {Graph Consulting},
  howpublished = {\url{https://graphsandnetworks.com/the-cora-dataset/}},
  note         = {Accessed: 2025-11-06}
}

@inproceedings{giles1998citeseer,
  title     = {CiteSeer: An automatic citation indexing system},
  author    = {Giles, C Lee and Bollacker, Kurt D and Lawrence, Steve},
  booktitle = {Proceedings of the third ACM conference on Digital libraries},
  pages     = {89--98},
  year      = {1998}
}

@misc{snapnets,
  author       = {Jure Leskovec and Andrej Krevl},
  title        = {{SNAP Datasets}: {Stanford} Large Network Dataset Collection},
  howpublished = {\url{http://snap.stanford.edu/data}},
  month        = jun,
  year         = {2014}
}

@article{zitnik2017ohmnet,
  title   = {Predicting multicellular function through multi-layer tissue networks},
  author  = {Zitnik, Marinka and Leskovec, Jure},
  journal = {Bioinformatics},
  year    = {2017},
  url     = {https://arxiv.org/abs/1707.04638}
}

@inproceedings{velickovic2018graph,
  title     = {Graph Attention Networks},
  author    = {Velickovic, Petar and Cucurull, Guillem and Casanova, Arantxa and Romero, Adriana and Lio, Pietro and Bengio, Yoshua},
  booktitle = {International Conference on Learning Representations},
  year      = {2018},
  url       = {https://arxiv.org/abs/1710.10903}
}
\clearpage

\appendix
\maketitlesupplementary
\setcounter{page}{1}
\raggedbottom
\setlength{\textfloatsep}{8pt plus 1pt minus 2pt}
\setlength{\dbltextfloatsep}{8pt plus 1pt minus 2pt}
\setlength{\floatsep}{6pt plus 1pt minus 2pt}
\setlength{\intextsep}{6pt plus 1pt minus 2pt}
\renewcommand{\topfraction}{0.92}
\renewcommand{\dbltopfraction}{0.92}
\renewcommand{\textfraction}{0.05}
\renewcommand{\floatpagefraction}{0.85}
\renewcommand{\dblfloatpagefraction}{0.85}

\definecolor{codeback}{rgb}{0.95,0.95,0.96}
\definecolor{codegreen}{HTML}{3CB371}  
\definecolor{codeblue}{HTML}{1E90FF}   
\definecolor{codeorange}{HTML}{E67E22} 
\definecolor{codegray}{rgb}{0.5,0.5,0.5}
\definecolor{codepurple}{rgb}{0.58,0,0.82}

\definecolor{bg_nodes}{HTML}{F9F2E7}   
\definecolor{bg_edges}{HTML}{E6F2FF}   
\definecolor{bg_degree}{HTML}{E6FFE6}  
\definecolor{bg_path}{HTML}{FFF7E6}    

\lstdefinestyle{defstyle}{
    backgroundcolor=\color{codeback},   
    commentstyle=\color{codegreen},
    keywordstyle=\color{codeblue}\bfseries,
    numberstyle=\tiny\color{codegray},
    stringstyle=\color{codeorange},
    basicstyle=\ttfamily\scriptsize, 
    breakatwhitespace=false,         
    breaklines=true,                 
    captionpos=b,                    
    keepspaces=true,                 
    numbers=left,                    
    numbersep=4pt,                  
    showspaces=false,                
    showstringspaces=false,
    showtabs=false,                  
    tabsize=2,
    frame=single,
    rulecolor=\color{black!10},
    xleftmargin=0.5em,
    xrightmargin=0.5em,
    morecomment=[l]{\#},
    morekeywords={VERSION, DESIGN, UNITS, DISTANCE, MICRONS, DIEAREA, ROW, STEP, DO, BY, LAYER, TRACKS, SPECIALNETS, END, COMPONENTS, PINS, NETS, VIAS, GCELLGRID, PLACED, FIXED, SOURCE, DIST, USE, SIGNAL, POWER, GROUND, ROUTED, SHAPE, STRIPE, NEW, DIRECTION, OUTPUT, INPUT, PORT, NET, FN, N, FS},
    sensitive=true
}

\lstdefinestyle{pythonstyle}{
    backgroundcolor=\color{codeback},   
    commentstyle=\color{codegreen},
    keywordstyle=\color{codeblue}\bfseries,
    numberstyle=\tiny\color{codegray},
    stringstyle=\color{codeorange},
    basicstyle=\ttfamily\scriptsize,       
    breakatwhitespace=false,         
    breaklines=true,                 
    captionpos=b,                    
    keepspaces=true,                 
    numbers=left,                    
    numbersep=4pt,                  
    showspaces=false,                
    showstringspaces=false,
    showtabs=false,                  
    tabsize=4,
    frame=single,
    language=Python,                       
    rulecolor=\color{black!10},
    xleftmargin=0.5em,
    xrightmargin=0.5em
}

This supplementary material provides comprehensive details, implementation specifics, and additional experimental analyses to support the main manuscript. The document is organized as follows: 
\textbf{Section~\ref{sec:usage}} offers a practical guide and API examples for utilizing the R2G benchmark suite.
\textbf{Section~\ref{sec:graph_datasets}} categorizes existing graph datasets to contextualize the unique contribution of the R2G benchmark. 
\textbf{Section~\ref{sec:def}} elaborates on the data processing pipeline, specifically the DEF parser used to translate physical layouts into graph structures. 
\textbf{Section~\ref{sec:stat}} presents granular statistics across the multi-view graph representations. 
\textbf{Section~\ref{sec:hp_tune}} provides an in-depth look at the experimental setup, detailing hyperparameter tuning strategies.
Finally, \textbf{Section~\ref{sec:gnn_head}} presents ablation studies on GNN and head depths.

\section{Usage}
\label{sec:usage}

R2G provides a reproducible pipeline that converts OpenROAD DEF files into multi-view circuit graphs and supports standardized GNN training and evaluation. The workflow consists of two main stages: (1) graph generation from physical-design files and (2) model training and evaluation on the generated datasets.

\subsection{Data Generation Pipeline}

R2G converts physical design outputs into graph datasets through a multi-stage pipeline. Starting from DEF files generated by the OpenROAD flow, the framework constructs typed circuit graphs and produces datasets compatible with PyTorch Geometric.

The pipeline consists of three steps:

\begin{itemize}
\item \textbf{Heterograph generation:} DEF files are parsed to construct typed heterogeneous graphs containing gates, nets, pins, and IO nodes. Multiple graph views (B--F) can be generated using the provided scripts in \texttt{data\_pipeline/heterograph\_generation/}.

\item \textbf{Homograph conversion:} The heterogeneous graphs are converted into homogeneous graphs for efficient GNN training using the converters in \texttt{data\_pipeline/homograph\_conversion/}.

\item \textbf{Graph merging:} Graphs from multiple designs are merged into a single dataset to enable large-scale training. The resulting merged graphs are stored as PyTorch \texttt{.pt} files and serve as the final benchmark datasets.
\end{itemize}

This pipeline converts DEF-based physical layouts into standardized circuit graphs suitable for machine learning tasks such as placement and routing prediction.

\subsection{Model Training}

The generated datasets can be used directly for GNN training. R2G provides training scripts for prediction tasks at different supervision granularities.

\begin{itemize}
\item \textbf{Node-centric configurations:} Training scripts are located in the \texttt{gnn-node/} directory. In the released baselines, they are used for placement-oriented targets such as HPWL.

\item \textbf{Edge-centric configurations:} Training scripts are located in the \texttt{gnn-edge/} directory. In the released baselines, they are used for routing-oriented targets such as wire-length prediction.
\end{itemize}

Each training pipeline includes dataset loading, feature encoding, neighbor sampling, model initialization, and evaluation. The framework supports multiple GNN architectures including GINE, ResGatedGCN, and GAT.

\subsection{Training Example}

A typical node-centric configuration for the placement task is shown below:

\begin{lstlisting}[style=pythonstyle, caption={Example training command for a node-centric placement configuration.}]
cd gnn-node

python main.py \
    --dataset place_B_homograph \
    --task_level node \
    --task regression \
    --model gine \
    --num_gnn_layers 4 \
    --num_head_layers 4 \
    --hid_dim 256 \
    --epochs 100 \
    --gpu 0
\end{lstlisting}

Similarly, routing tasks can be trained using the scripts in \texttt{gnn-edge/}.

\subsection{Evaluation}

During training, the framework automatically reports standard regression metrics including Mean Absolute Error (MAE), Root Mean Squared Error (RMSE), and the coefficient of determination ($R^2$). The best-performing model on the validation set is saved and evaluated on the test set.

Additional analysis tools are provided to visualize prediction quality, including scatter plots and label distribution statistics.

\section{Overview of existing graph datasets}
\label{sec:graph_datasets}
\begin{table}[tb]
  \centering
  \caption{Overview of existing graph datasets. Tasks include Classification (C), Regression (R), and Link prediction (L).}
  \label{tab:graph_datasets}
  \resizebox{\linewidth}{!}
  {
    \begin{tabular}{cc|ccc cccc}
      \toprule
      \multicolumn{2}{c}{\multirow{2}{*}{\textbf{Benchmark Suite}}} & 
      
      \multicolumn{1}{c}{\multirow{2}{*}{\textbf{Category}}} & 
      
      \multicolumn{1}{c}{\multirow{2}{*}{\textbf{Level}}} & 
      
      \multicolumn{1}{c}{\multirow{2}{*}{\textbf{Task}}}& 
      
      \multicolumn{3}{c}{\textbf{Statistics}} & 
      
      \multicolumn{1}{c}{\multirow{2}{*}{\textbf{Metric}}} \\
      
      \cmidrule{6-8} 

       \multicolumn{2}{c}{} & & & & \textbf{\#Node/G} & \textbf{\#Edge/G} & \textbf{\#Degree/G} & \\
      \midrule
      
      \multirow{4}{*}{\textbf{OGB}} 
      & \cellcolor[HTML]{F9F2E7}ogbn-products & \cellcolor[HTML]{F9F2E7}Product & \cellcolor[HTML]{F9F2E7}node & \cellcolor[HTML]{F9F2E7}C & \cellcolor[HTML]{F9F2E7}2,449,029 & \cellcolor[HTML]{F9F2E7}61,859,140 & \cellcolor[HTML]{F9F2E7}50.5 & \cellcolor[HTML]{F9F2E7}Accuracy \\
      & \cellcolor[HTML]{F9F2E7}ogbn-mag & \cellcolor[HTML]{F9F2E7}Academic & \cellcolor[HTML]{F9F2E7}node & \cellcolor[HTML]{F9F2E7}C & \cellcolor[HTML]{F9F2E7}1,939,743 & \cellcolor[HTML]{F9F2E7}21,111,007 & \cellcolor[HTML]{F9F2E7}21.77 & \cellcolor[HTML]{F9F2E7}Accuracy \\
      & \cellcolor[HTML]{F9F2E7}ogbl-ppa & \cellcolor[HTML]{F9F2E7}Bio & \cellcolor[HTML]{F9F2E7}edge & \cellcolor[HTML]{F9F2E7}L & \cellcolor[HTML]{F9F2E7}576,289 & \cellcolor[HTML]{F9F2E7}30,326,273 & \cellcolor[HTML]{F9F2E7}105.2 & \cellcolor[HTML]{F9F2E7}Hits@100 \\
      & \cellcolor[HTML]{F9F2E7}ogbl-citation2 & \cellcolor[HTML]{F9F2E7}Citation & \cellcolor[HTML]{F9F2E7}edge & \cellcolor[HTML]{F9F2E7}L & \cellcolor[HTML]{F9F2E7}2,927,963 & \cellcolor[HTML]{F9F2E7}30,561,187 & \cellcolor[HTML]{F9F2E7}20.9 & \cellcolor[HTML]{F9F2E7}MRR \\
      \midrule
      \multirow{5}{*}{\textbf{TU}} 
      & \cellcolor[HTML]{E6F2FF}QM9 & \cellcolor[HTML]{E6F2FF}Molecules & \cellcolor[HTML]{E6F2FF}node & \cellcolor[HTML]{E6F2FF}R & \cellcolor[HTML]{E6F2FF}18 & \cellcolor[HTML]{E6F2FF}19 & \cellcolor[HTML]{E6F2FF}2.07 & \cellcolor[HTML]{E6F2FF}MAE \\
      & \cellcolor[HTML]{E6F2FF}ZINC & \cellcolor[HTML]{E6F2FF}Molecules & \cellcolor[HTML]{E6F2FF}graph & \cellcolor[HTML]{E6F2FF}R & \cellcolor[HTML]{E6F2FF}23 & \cellcolor[HTML]{E6F2FF}25 & \cellcolor[HTML]{E6F2FF}2.14 & \cellcolor[HTML]{E6F2FF}MAE \\
      & \cellcolor[HTML]{E6F2FF}PROTEINS & \cellcolor[HTML]{E6F2FF}Bio & \cellcolor[HTML]{E6F2FF}graph & \cellcolor[HTML]{E6F2FF}C & \cellcolor[HTML]{E6F2FF}39 & \cellcolor[HTML]{E6F2FF}73 & \cellcolor[HTML]{E6F2FF}3.73 & \cellcolor[HTML]{E6F2FF}Accuracy \\
      & \cellcolor[HTML]{E6F2FF}reddit & \cellcolor[HTML]{E6F2FF}Social & \cellcolor[HTML]{E6F2FF}graph & \cellcolor[HTML]{E6F2FF}C & \cellcolor[HTML]{E6F2FF}24 & \cellcolor[HTML]{E6F2FF}25 & \cellcolor[HTML]{E6F2FF}2.03 & \cellcolor[HTML]{E6F2FF}F1-Score \\
      & \cellcolor[HTML]{E6F2FF}COLLAB & \cellcolor[HTML]{E6F2FF}Social & \cellcolor[HTML]{E6F2FF}graph & \cellcolor[HTML]{E6F2FF}C & \cellcolor[HTML]{E6F2FF}74 & \cellcolor[HTML]{E6F2FF}2,458 & \cellcolor[HTML]{E6F2FF}37.39 & \cellcolor[HTML]{E6F2FF}F1-Score \\
      \midrule
      \multirow{5}{*}{\textbf{SNAP}} 
      & \cellcolor[HTML]{E6FFE6}ego-Twitter & \cellcolor[HTML]{E6FFE6}Social & \cellcolor[HTML]{E6FFE6}node & \cellcolor[HTML]{E6FFE6}C & \cellcolor[HTML]{E6FFE6}81306 & \cellcolor[HTML]{E6FFE6}1768149 & \cellcolor[HTML]{E6FFE6}43.49 & \cellcolor[HTML]{E6FFE6}F1-Score \\
      & \cellcolor[HTML]{E6FFE6}com-Youtube & \cellcolor[HTML]{E6FFE6}Social & \cellcolor[HTML]{E6FFE6}node & \cellcolor[HTML]{E6FFE6}C & \cellcolor[HTML]{E6FFE6}1134890 & \cellcolor[HTML]{E6FFE6}2987624 & \cellcolor[HTML]{E6FFE6}5.27 & \cellcolor[HTML]{E6FFE6}F1-Score \\
      & \cellcolor[HTML]{E6FFE6}cit-Patents & \cellcolor[HTML]{E6FFE6}Citation & \cellcolor[HTML]{E6FFE6}node & \cellcolor[HTML]{E6FFE6}C & \cellcolor[HTML]{E6FFE6}3774768 & \cellcolor[HTML]{E6FFE6}16518948 & \cellcolor[HTML]{E6FFE6}8.75 & \cellcolor[HTML]{E6FFE6}F1-Score \\
      & \cellcolor[HTML]{E6FFE6}web-Google & \cellcolor[HTML]{E6FFE6}Web & \cellcolor[HTML]{E6FFE6}node & \cellcolor[HTML]{E6FFE6}C & \cellcolor[HTML]{E6FFE6}875713 & \cellcolor[HTML]{E6FFE6}5105039 & \cellcolor[HTML]{E6FFE6}11.66 & \cellcolor[HTML]{E6FFE6}F1-Score \\
      & \cellcolor[HTML]{E6FFE6}amazon-meta & \cellcolor[HTML]{E6FFE6}Product & \cellcolor[HTML]{E6FFE6}node & \cellcolor[HTML]{E6FFE6}C & \cellcolor[HTML]{E6FFE6}548552 & \cellcolor[HTML]{E6FFE6}1788725 & \cellcolor[HTML]{E6FFE6}6.52 & \cellcolor[HTML]{E6FFE6}F1-Score \\
      \midrule
      \multirow{5}{*}{\textbf{R2G}} 
      & \cellcolor[HTML]{FFF7E6}B\_graph & \cellcolor[HTML]{FFF7E6}Circuit Design & \cellcolor[HTML]{FFF7E6}node & \cellcolor[HTML]{FFF7E6}R & \cellcolor[HTML]{FFF7E6}124408 & \cellcolor[HTML]{FFF7E6}149287 & \cellcolor[HTML]{FFF7E6}2.4 & \cellcolor[HTML]{FFF7E6}MAE \\
      & \cellcolor[HTML]{FFF7E6}C\_graph & \cellcolor[HTML]{FFF7E6}Circuit Design & \cellcolor[HTML]{FFF7E6}node & \cellcolor[HTML]{FFF7E6}R & \cellcolor[HTML]{FFF7E6}49921 & \cellcolor[HTML]{FFF7E6}110166 & \cellcolor[HTML]{FFF7E6}4.41 & \cellcolor[HTML]{FFF7E6}MAE \\
      & \cellcolor[HTML]{FFF7E6}D\_graph & \cellcolor[HTML]{FFF7E6}Circuit Design & \cellcolor[HTML]{FFF7E6}edge & \cellcolor[HTML]{FFF7E6}R/C & \cellcolor[HTML]{FFF7E6}89640 & \cellcolor[HTML]{FFF7E6}107689 & \cellcolor[HTML]{FFF7E6}2.4 & \cellcolor[HTML]{FFF7E6}MAE/F1-Score \\
      & \cellcolor[HTML]{FFF7E6}E\_graph & \cellcolor[HTML]{FFF7E6}Circuit Design & \cellcolor[HTML]{FFF7E6}edge & \cellcolor[HTML]{FFF7E6}R/C & \cellcolor[HTML]{FFF7E6}64975 & \cellcolor[HTML]{FFF7E6}78789 & \cellcolor[HTML]{FFF7E6}2.43 & \cellcolor[HTML]{FFF7E6}MAE/F1-Score \\
      & \cellcolor[HTML]{FFF7E6}F\_graph & \cellcolor[HTML]{FFF7E6}Circuit Design & \cellcolor[HTML]{FFF7E6}edge & \cellcolor[HTML]{FFF7E6}R/C & \cellcolor[HTML]{FFF7E6}25061 & \cellcolor[HTML]{FFF7E6}46597 & \cellcolor[HTML]{FFF7E6}3.72 & \cellcolor[HTML]{FFF7E6}MAE/F1-Score \\
      \bottomrule
    \end{tabular}
    }
\end{table}
Benchmarks and tasks. We situate widely used graph suites alongside our circuit-graph benchmark, as summarized in \autoref{tab:graph_datasets}. OGB~\cite{hu2020open} targets large-scale node, link, and graph tasks with unified splits and metrics; TUDataset~\cite{morris2020tudataset} aggregates small-to-mid collections (molecules, bio, social) for graph-level classification and regression; SNAP~\cite{snapnets} curates massive social, web, and citation networks for node classification and link prediction. These domain-agnostic, predominantly single-view resources lack typed heterogeneity, multi-terminal hyperedges, and geometry-aware attributes required for physical design.

R2G differs fundamentally. It releases typed, heterogeneous circuit graphs extracted from DEF, with stage-aware, multi-view representations and \emph{information parity}. Targets are attached at the entity resolution induced by each view, with domain metrics aligned to EDA objectives in placement and routing. Graphs are industrial yet tractable, capturing net topology, pin-level connectivity, and macro-to-cell context. With scalable loaders, unified splits, metrics, and reproducible baselines, R2G enables fair cross-view comparison and cleanly decouples representation from modeling across late physical-design stages.

\section{DEF Parser}
\label{sec:def}

To bridge the gap between physical layout implementation and graph-based learning, the R2G framework incorporates a specialized DEF parser. This component is responsible for extracting semantic and geometric information from standard Design Exchange Format (DEF) files, converting them into the structured graph representations used in our benchmark. In this section, we first detail the structural evolution of DEF files throughout the design flow to clarify the data availability at each stage. Subsequently, we outline the algorithmic procedure used to translate these raw descriptions into heterogeneous graphs.

\subsection{DEF File Description}

The Design Exchange Format (DEF) serves as the standard for exchanging physical layout information. Its content evolves significantly through the physical design stages:

\begin{itemize}
    \item \textbf{Floorplan Stage:} This stage establishes the ``geometric stage'' and power distribution. The DEF file defines the design bounds (\texttt{DIEAREA}), standard cell rows (\texttt{ROW}), and routing tracks (\texttt{TRACKS}). Power networks (\texttt{SPECIALNETS}) are often generated here (e.g., VDD/VSS stripes). Logical connections (\texttt{NETS}) exist but lack physical geometry, and components generally lack fixed coordinates unless manually locked.
    
    \item \textbf{Placement Stage:} Specific physical coordinates are assigned. Standard cells in the \texttt{COMPONENTS} section are updated with \texttt{+ PLACED (x y)} coordinates and orientations. Similarly, I/O pins in the \texttt{PINS} section are assigned specific metal layers and physical locations (\texttt{+ PORT + LAYER}). Signal nets remain logical connections without detailed routing segments.
    
    \item \textbf{Routing Stage:} The layout is finalized with detailed interconnects. A global routing grid (\texttt{GCELLGRID}) is often added. The \texttt{NETS} section is populated with detailed physical paths, including wire segments, turns, and cross-layer vias using the \texttt{+ ROUTED ... NEW} syntax. The component count may increase due to the insertion of filler cells or buffers.
\end{itemize}

Listing~\ref{lst:def_evolution} illustrates a routed DEF file structure, highlighting the information accumulated from these stages.
\begin{figure}[tb]
\centering
\begin{lstlisting}[style=defstyle, caption={A DEF file example showing the accumulation of data from Floorplan to Routing.}, label={lst:def_evolution}]
DESIGN ac97_top ;
UNITS DISTANCE MICRONS 2000 ;
DIEAREA ( 0 0 ) ( 434390 434390 ) ;

# [Floorplan] Rows, Tracks, and Power Grid defined
ROW ROW_0 FreePDK45_38x28_10R_NP_162NW_34O 2280 2800 N DO 1131 BY 1 STEP 380 0 ;
TRACKS X 190 DO 1143 STEP 380 LAYER metal1 ;
SPECIALNETS 2 ;
    - VDD ( * VDD ) + USE POWER
      + ROUTED metal7 2800 + SHAPE STRIPE ( 61800 396800 ) ( 398760 396800 ) ;
END SPECIALNETS

# [Routing] Global routing grid added
GCELLGRID X 0 DO 103 STEP 4200 ;

# [Placement] Coords added (+ PLACED); [Routing] Count increases
COMPONENTS 11178 ;
    - _10221_ CLKBUF_X2 + PLACED ( 169860 109200 ) FN ;
    - _10222_ INV_X1    + PLACED ( 165300 109200 ) FN ;
END COMPONENTS

# [Placement] Pin physical details (+ PORT + LAYER) added
PINS 132 ;
    - ac97_reset_pad_o_ + NET ac97_reset_pad_o_ + DIRECTION OUTPUT + USE SIGNAL
      + PORT
        + LAYER metal6 ( -140 -140 ) ( 140 140 )
        + PLACED ( 127870 140 ) N ;
END PINS

# [Routing] Detailed geometry segments (+ ROUTED ... NEW) added
NETS 12959 ;
    - _00000_ ( u10.dout[0]$_DFF_P_ D ) ( _11175_ ZN ) + USE SIGNAL
      + ROUTED metal2 ( 91770 88060 ) ( * 96600 )
      NEW    metal2 ( 91770 96600 ) ( 92150 * )
      NEW    metal1 ( 91770 88060 ) via1_4 ;
END NETS

END DESIGN
\end{lstlisting}
\end{figure}

\subsection{DEF-to-Graph Conversion}
\label{subsec:def2g}
\begin{algorithm}[tb]
\caption{DEF-to-Graph Conversion Process}
\label{alg:def2graph}
\small 
\begin{algorithmic}[1] 
\REQUIRE DEF file $\mathcal{D}$
\ENSURE Heterogeneous Graph $\mathcal{G} = (\mathcal{V}, \mathcal{E})$

\STATE \textbf{Initialization:} $\mathcal{V} \leftarrow \emptyset, \mathcal{E} \leftarrow \emptyset$
\STATE $\mathcal{D}_{data} \leftarrow \text{ParseDEF}(\mathcal{D})$

\vspace{0.3em}
\STATE \COMMENT{\textbf{Phase 1: Node Construction}}
\FORALL{$comp \in \mathcal{D}_{data}.components$}
    \STATE $v_{gate} \leftarrow \text{CreateGateNode}(comp)$
    \STATE $\mathcal{V}.add(v_{gate})$
\ENDFOR

\FORALL{$net \in \mathcal{D}_{data}.nets$}
    \STATE $v_{net} \leftarrow \text{CreateNetNode}(net)$
    \STATE $\mathcal{V}.add(v_{net})$
\ENDFOR

\FORALL{$pin \in \mathcal{D}_{data}.pins$}
    \STATE $v_{io} \leftarrow \text{CreateIOPinNode}(pin)$
    \STATE $\mathcal{V}.add(v_{io})$
\ENDFOR

\FORALL{$int\_pin \in \mathcal{D}_{data}.internal\_pins$}
    \STATE $v_{pin} \leftarrow \text{CreatePinNode}(int\_pin)$
    \STATE $\mathcal{V}.add(v_{pin})$
\ENDFOR

\vspace{0.3em}
\STATE \COMMENT{\textbf{Phase 2: Edge Construction}}
\FORALL{$net \in \mathcal{D}_{data}.nets$}
    \FORALL{$(comp, pin) \in net.connections$}
        \IF{$comp == \text{`PIN'}$}
            \STATE $\mathcal{E}.add((v_{io}, v_{net}))$ \COMMENT{Edge: IO Pin $\leftrightarrow$ Net}
        \ELSE
            \STATE $\mathcal{E}.add((v_{pin}, v_{net}))$ \COMMENT{Edge: Internal Pin $\leftrightarrow$ Net}
        \ENDIF
    \ENDFOR
\ENDFOR

\FORALL{$comp \in \mathcal{D}_{data}.components$}
    \FORALL{$pin \in comp.pins$}
        \STATE $\mathcal{E}.add((v_{gate}, v_{pin}))$ \COMMENT{Edge: Gate $\leftrightarrow$ Pin}
    \ENDFOR
\ENDFOR

\vspace{0.3em}
\STATE \COMMENT{\textbf{Phase 3: Feature Encoding}}
\FORALL{$v \in \mathcal{V}$}
    \STATE $v.x \leftarrow \text{EncodeNodeFeatures}(v)$
\ENDFOR
\FORALL{$e \in \mathcal{E}$}
    \STATE $e.attr \leftarrow \text{EncodeEdgeFeatures}(e)$
\ENDFOR
\STATE $\mathcal{G}.global \leftarrow \text{ExtractGlobalFeatures}(\mathcal{D}_{data})$

\RETURN $\mathcal{G}$
\end{algorithmic}
\end{algorithm}
The conversion process from a raw DEF file to a structured graph representation is outlined in Algorithm~\ref{alg:def2graph}.

The DEF-to-Graph Conversion Algorithm transforms a Design Exchange Format (DEF) file, which describes the physical layout of an integrated circuit, into a heterogeneous graph representation suitable for machine learning and analysis tasks. This algorithm systematically converts circuit components into graph nodes and their interconnections into graph edges, creating a structured representation that preserves the topological and functional relationships within the circuit design.

The conversion process involves three main phases: Node Construction (creating nodes for gates, nets, and pins), Edge Construction (establishing connectivity between components), and Feature Encoding (adding numerical attributes to nodes and edges). The resulting heterogeneous graph captures both the structural hierarchy and physical properties of the circuit, enabling applications in optimization, verification, and AI-driven design automation.

\section{Statistics of Multi-Views}
\label{sec:stat}

\begin{table*}[t]
\centering
\caption{Comparison of graph statistics across graph views and design categories.}
\label{tab:view_stats}
\resizebox{\linewidth}{!}{%
\begin{tabular}{c ccccccccccccccccccccccc}
\toprule
\multicolumn{2}{c}{} & \multicolumn{5}{c}{\cellcolor[HTML]{F9F2E7}\textbf{\#nodes}} & \multicolumn{5}{c}{\cellcolor[HTML]{E6F2FF}\textbf{\#edges}} & \multicolumn{5}{c}{\cellcolor[HTML]{E6FFE6}\textbf{avg\_degree}} & \multicolumn{5}{c}{\cellcolor[HTML]{FFF7E6}\textbf{avg\_shortest path}} \\

\cmidrule(lr){3-7}  \cmidrule(lr){8-12} \cmidrule(lr){13-17} \cmidrule(lr){18-22}

\textbf{Cat.} & \textbf{RTL2G} & 
\cellcolor[HTML]{F9F2E7}\textbf{(b)} & \cellcolor[HTML]{F9F2E7}\textbf{(c)} & \cellcolor[HTML]{F9F2E7}\textbf{(d)} & \cellcolor[HTML]{F9F2E7}\textbf{(e)} & \cellcolor[HTML]{F9F2E7}\textbf{(f)} & 
\cellcolor[HTML]{E6F2FF}\textbf{(b)} & \cellcolor[HTML]{E6F2FF}\textbf{(c)} & \cellcolor[HTML]{E6F2FF}\textbf{(d)} & \cellcolor[HTML]{E6F2FF}\textbf{(e)} & \cellcolor[HTML]{E6F2FF}\textbf{(f)} & 
\cellcolor[HTML]{E6FFE6}\textbf{(b)} & \cellcolor[HTML]{E6FFE6}\textbf{(c)} & \cellcolor[HTML]{E6FFE6}\textbf{(d)} & \cellcolor[HTML]{E6FFE6}\textbf{(e)} & \cellcolor[HTML]{E6FFE6}\textbf{(f)} & 
\cellcolor[HTML]{FFF7E6}\textbf{(b)} & \cellcolor[HTML]{FFF7E6}\textbf{(c)} & \cellcolor[HTML]{FFF7E6}\textbf{(d)} & \cellcolor[HTML]{FFF7E6}\textbf{(e)} & \cellcolor[HTML]{FFF7E6}\textbf{(f)} \\ 
\midrule

\multirow{4}{*}{\rotatebox[origin=c]{90}{\textbf{A/V ctrl.}}} 
& ss\_pcm & 
\cellcolor[HTML]{F9F2E7}2.44k & \cellcolor[HTML]{F9F2E7}1.01k & \cellcolor[HTML]{F9F2E7}1.89k & \cellcolor[HTML]{F9F2E7}1.46k & \cellcolor[HTML]{F9F2E7}0.46k & 
\cellcolor[HTML]{E6F2FF}2.88k & \cellcolor[HTML]{E6F2FF}2.02k & \cellcolor[HTML]{E6F2FF}2.36k & \cellcolor[HTML]{E6F2FF}1.66k & \cellcolor[HTML]{E6F2FF}1.01k & 
\cellcolor[HTML]{E6FFE6}2.37 & \cellcolor[HTML]{E6FFE6}4.01 & \cellcolor[HTML]{E6FFE6}2.49 & \cellcolor[HTML]{E6FFE6}2.28 & \cellcolor[HTML]{E6FFE6}4.35 & 
\cellcolor[HTML]{FFF7E6}12.07 & \cellcolor[HTML]{FFF7E6}3.00 & \cellcolor[HTML]{FFF7E6}9.81 & \cellcolor[HTML]{FFF7E6}12.69 & \cellcolor[HTML]{FFF7E6}4.05 \\
& ac97\_ctrl & 
\cellcolor[HTML]{F9F2E7}58.22k & \cellcolor[HTML]{F9F2E7}22.99k & \cellcolor[HTML]{F9F2E7}40.86k & \cellcolor[HTML]{F9F2E7}30.63k & \cellcolor[HTML]{F9F2E7}10.36k & 
\cellcolor[HTML]{E6F2FF}70.60k & \cellcolor[HTML]{E6F2FF}47.20k & \cellcolor[HTML]{E6F2FF}52.19k & \cellcolor[HTML]{E6F2FF}36.83k & \cellcolor[HTML]{E6F2FF}23.50k & 
\cellcolor[HTML]{E6FFE6}2.43 & \cellcolor[HTML]{E6FFE6}4.11 & \cellcolor[HTML]{E6FFE6}2.55 & \cellcolor[HTML]{E6FFE6}2.40 & \cellcolor[HTML]{E6FFE6}4.54 & 
\cellcolor[HTML]{FFF7E6}13.41 & \cellcolor[HTML]{FFF7E6}3.00 & \cellcolor[HTML]{FFF7E6}13.76 & \cellcolor[HTML]{FFF7E6}21.20 & \cellcolor[HTML]{FFF7E6}5.65 \\
& vga\_lcd & 
\cellcolor[HTML]{F9F2E7}531.33k & \cellcolor[HTML]{F9F2E7}206.01k & \cellcolor[HTML]{F9F2E7}369.72k & \cellcolor[HTML]{F9F2E7}275.41k & \cellcolor[HTML]{F9F2E7}94.51k & 
\cellcolor[HTML]{E6F2FF}650.84k & \cellcolor[HTML]{E6F2FF}445.88k & \cellcolor[HTML]{E6F2FF}455.43k & \cellcolor[HTML]{E6F2FF}325.55k & \cellcolor[HTML]{E6F2FF}197.12k & 
\cellcolor[HTML]{E6FFE6}2.45 & \cellcolor[HTML]{E6FFE6}4.33 & \cellcolor[HTML]{E6FFE6}2.46 & \cellcolor[HTML]{E6FFE6}2.36 & \cellcolor[HTML]{E6FFE6}4.17 & 
\cellcolor[HTML]{FFF7E6}14.33 & \cellcolor[HTML]{FFF7E6}3.02 & \cellcolor[HTML]{FFF7E6}24.37 & \cellcolor[HTML]{FFF7E6}31.42 & \cellcolor[HTML]{FFF7E6}8.00 \\
\cmidrule{2-22} 
 &\textbf{Average} & 
\cellcolor[HTML]{F9F2E7}197.33k & \cellcolor[HTML]{F9F2E7}76.67k & \cellcolor[HTML]{F9F2E7}137.49k & \cellcolor[HTML]{F9F2E7}102.50k & \cellcolor[HTML]{F9F2E7}35.11k & 
\cellcolor[HTML]{E6F2FF}241.44k & \cellcolor[HTML]{E6F2FF}165.03k & \cellcolor[HTML]{E6F2FF}170.00k & \cellcolor[HTML]{E6F2FF}121.35k & \cellcolor[HTML]{E6F2FF}74.08k & 
\cellcolor[HTML]{E6FFE6}2.42 & \cellcolor[HTML]{E6FFE6}4.15 & \cellcolor[HTML]{E6FFE6}2.50 & \cellcolor[HTML]{E6FFE6}2.35 & \cellcolor[HTML]{E6FFE6}4.35 & 
\cellcolor[HTML]{FFF7E6}13.27 & \cellcolor[HTML]{FFF7E6}3.01 & \cellcolor[HTML]{FFF7E6}15.98 & \cellcolor[HTML]{FFF7E6}21.77 & \cellcolor[HTML]{FFF7E6}5.90 \\
\midrule

\multirow{7}{*}{\rotatebox[origin=c]{90}{\textbf{Crypto Core}}} 
& des3\_area & 
\cellcolor[HTML]{F9F2E7}9.78k & \cellcolor[HTML]{F9F2E7}3.65k & \cellcolor[HTML]{F9F2E7}7.94k & \cellcolor[HTML]{F9F2E7}6.33k & \cellcolor[HTML]{F9F2E7}1.80k & 
\cellcolor[HTML]{E6F2FF}12.46k & \cellcolor[HTML]{E6F2FF}9.29k & \cellcolor[HTML]{E6F2FF}11.17k & \cellcolor[HTML]{E6F2FF}9.41k & \cellcolor[HTML]{E6F2FF}5.10k & 
\cellcolor[HTML]{E6FFE6}2.55 & \cellcolor[HTML]{E6FFE6}5.09 & \cellcolor[HTML]{E6FFE6}2.81 & \cellcolor[HTML]{E6FFE6}2.97 & \cellcolor[HTML]{E6FFE6}5.66 & 
\cellcolor[HTML]{FFF7E6}18.16 & \cellcolor[HTML]{FFF7E6}3.28 & \cellcolor[HTML]{FFF7E6}15.43 & \cellcolor[HTML]{FFF7E6}12.42 & \cellcolor[HTML]{FFF7E6}6.13 \\
& systemcdes & 
\cellcolor[HTML]{F9F2E7}14.17k & \cellcolor[HTML]{F9F2E7}5.76k & \cellcolor[HTML]{F9F2E7}10.73k & \cellcolor[HTML]{F9F2E7}8.14k & \cellcolor[HTML]{F9F2E7}2.79k & 
\cellcolor[HTML]{E6F2FF}17.02k & \cellcolor[HTML]{E6F2FF}12.84k & \cellcolor[HTML]{E6F2FF}13.48k & \cellcolor[HTML]{E6F2FF}10.13k & \cellcolor[HTML]{E6F2FF}5.73k & 
\cellcolor[HTML]{E6FFE6}2.40 & \cellcolor[HTML]{E6FFE6}4.46 & \cellcolor[HTML]{E6FFE6}2.51 & \cellcolor[HTML]{E6FFE6}2.49 & \cellcolor[HTML]{E6FFE6}4.11 & 
\cellcolor[HTML]{FFF7E6}20.60 & \cellcolor[HTML]{FFF7E6}3.18 & \cellcolor[HTML]{FFF7E6}19.62 & \cellcolor[HTML]{FFF7E6}17.26 & \cellcolor[HTML]{FFF7E6}7.20 \\
& systemcaes & 
\cellcolor[HTML]{F9F2E7}41.08k & \cellcolor[HTML]{F9F2E7}16.13k & \cellcolor[HTML]{F9F2E7}31.74k & \cellcolor[HTML]{F9F2E7}24.34k & \cellcolor[HTML]{F9F2E7}7.78k & 
\cellcolor[HTML]{E6F2FF}50.29k & \cellcolor[HTML]{E6F2FF}36.81k & \cellcolor[HTML]{E6F2FF}40.30k & \cellcolor[HTML]{E6F2FF}30.46k & \cellcolor[HTML]{E6F2FF}16.83k & 
\cellcolor[HTML]{E6FFE6}2.45 & \cellcolor[HTML]{E6FFE6}4.56 & \cellcolor[HTML]{E6FFE6}2.54 & \cellcolor[HTML]{E6FFE6}2.50 & \cellcolor[HTML]{E6FFE6}4.33 & 
\cellcolor[HTML]{FFF7E6}18.02 & \cellcolor[HTML]{FFF7E6}3.16 & \cellcolor[HTML]{FFF7E6}13.91 & \cellcolor[HTML]{FFF7E6}18.13 & \cellcolor[HTML]{FFF7E6}6.20 \\
& sha256 & 
\cellcolor[HTML]{F9F2E7}62.83k & \cellcolor[HTML]{F9F2E7}25.06k & \cellcolor[HTML]{F9F2E7}47.57k & \cellcolor[HTML]{F9F2E7}36.52k & \cellcolor[HTML]{F9F2E7}11.83k & 
\cellcolor[HTML]{E6F2FF}76.32k & \cellcolor[HTML]{E6F2FF}56.36k & \cellcolor[HTML]{E6F2FF}59.51k & \cellcolor[HTML]{E6F2FF}45.47k & \cellcolor[HTML]{E6F2FF}24.70k & 
\cellcolor[HTML]{E6FFE6}2.43 & \cellcolor[HTML]{E6FFE6}4.50 & \cellcolor[HTML]{E6FFE6}2.50 & \cellcolor[HTML]{E6FFE6}2.49 & \cellcolor[HTML]{E6FFE6}4.18 & 
\cellcolor[HTML]{FFF7E6}18.66 & \cellcolor[HTML]{FFF7E6}2.88 & \cellcolor[HTML]{FFF7E6}16.43 & \cellcolor[HTML]{FFF7E6}17.12 & \cellcolor[HTML]{FFF7E6}6.75 \\
& aes\_secworks & 
\cellcolor[HTML]{F9F2E7}128.40k & \cellcolor[HTML]{F9F2E7}46.60k & \cellcolor[HTML]{F9F2E7}96.64k & \cellcolor[HTML]{F9F2E7}75.04k & \cellcolor[HTML]{F9F2E7}22.12k & 
\cellcolor[HTML]{E6F2FF}164.11k & \cellcolor[HTML]{E6F2FF}113.16k & \cellcolor[HTML]{E6F2FF}128.92k & \cellcolor[HTML]{E6F2FF}99.87k & \cellcolor[HTML]{E6F2FF}56.87k & 
\cellcolor[HTML]{E6FFE6}2.56 & \cellcolor[HTML]{E6FFE6}4.86 & \cellcolor[HTML]{E6FFE6}2.67 & \cellcolor[HTML]{E6FFE6}2.66 & \cellcolor[HTML]{E6FFE6}5.14 & 
\cellcolor[HTML]{FFF7E6}19.92 & \cellcolor[HTML]{FFF7E6}2.93 & \cellcolor[HTML]{FFF7E6}19.73 & \cellcolor[HTML]{FFF7E6}20.61 & \cellcolor[HTML]{FFF7E6}7.07 \\
& aes\_xcrypt & 
\cellcolor[HTML]{F9F2E7}165.76k & \cellcolor[HTML]{F9F2E7}58.42k & \cellcolor[HTML]{F9F2E7}135.12k & \cellcolor[HTML]{F9F2E7}107.08k & \cellcolor[HTML]{F9F2E7}28.43k & 
\cellcolor[HTML]{E6F2FF}215.07k & \cellcolor[HTML]{E6F2FF}156.35k & \cellcolor[HTML]{E6F2FF}188.34k & \cellcolor[HTML]{E6F2FF}156.69k & \cellcolor[HTML]{E6F2FF}81.88k & 
\cellcolor[HTML]{E6FFE6}2.60 & \cellcolor[HTML]{E6FFE6}5.35 & \cellcolor[HTML]{E6FFE6}2.79 & \cellcolor[HTML]{E6FFE6}2.93 & \cellcolor[HTML]{E6FFE6}5.76 & 
\cellcolor[HTML]{FFF7E6}20.68 & \cellcolor[HTML]{FFF7E6}2.97 & \cellcolor[HTML]{FFF7E6}17.39 & \cellcolor[HTML]{FFF7E6}16.57 & \cellcolor[HTML]{FFF7E6}7.17 \\
\cmidrule{2-22}
 &\textbf{Average} & 
\cellcolor[HTML]{F9F2E7}73.85k & \cellcolor[HTML]{F9F2E7}26.27k & \cellcolor[HTML]{F9F2E7}58.29k & \cellcolor[HTML]{F9F2E7}44.58k & \cellcolor[HTML]{F9F2E7}12.46k & 
\cellcolor[HTML]{E6F2FF}92.55k & \cellcolor[HTML]{E6F2FF}64.13k & \cellcolor[HTML]{E6F2FF}75.32k & \cellcolor[HTML]{E6F2FF}58.67k & \cellcolor[HTML]{E6F2FF}31.84k & 
\cellcolor[HTML]{E6FFE6}2.50 & \cellcolor[HTML]{E6FFE6}4.77 & \cellcolor[HTML]{E6FFE6}2.64 & \cellcolor[HTML]{E6FFE6}2.64 & \cellcolor[HTML]{E6FFE6}4.86 & 
\cellcolor[HTML]{FFF7E6}19.34 & \cellcolor[HTML]{FFF7E6}3.07 & \cellcolor[HTML]{FFF7E6}17.08 & \cellcolor[HTML]{FFF7E6}17.02 & \cellcolor[HTML]{FFF7E6}6.75 \\
\midrule

\multirow{8}{*}{\rotatebox[origin=c]{90}{\textbf{Processor}}} 
& tv80 & 
\cellcolor[HTML]{F9F2E7}31.14k & \cellcolor[HTML]{F9F2E7}11.44k & \cellcolor[HTML]{F9F2E7}24.91k & \cellcolor[HTML]{F9F2E7}19.47k & \cellcolor[HTML]{F9F2E7}5.51k & 
\cellcolor[HTML]{E6F2FF}39.45k & \cellcolor[HTML]{E6F2FF}29.34k & \cellcolor[HTML]{E6F2FF}34.25k & \cellcolor[HTML]{E6F2FF}28.48k & \cellcolor[HTML]{E6F2FF}15.14k & 
\cellcolor[HTML]{E6FFE6}2.53 & \cellcolor[HTML]{E6FFE6}5.13 & \cellcolor[HTML]{E6FFE6}2.75 & \cellcolor[HTML]{E6FFE6}2.93 & \cellcolor[HTML]{E6FFE6}5.50 & 
\cellcolor[HTML]{FFF7E6}16.71 & \cellcolor[HTML]{FFF7E6}3.10 & \cellcolor[HTML]{FFF7E6}14.48 & \cellcolor[HTML]{FFF7E6}13.95 & \cellcolor[HTML]{FFF7E6}5.94 \\
& tv80s & 
\cellcolor[HTML]{F9F2E7}37.24k & \cellcolor[HTML]{F9F2E7}14.39k & \cellcolor[HTML]{F9F2E7}28.88k & \cellcolor[HTML]{F9F2E7}21.96k & \cellcolor[HTML]{F9F2E7}6.97k & 
\cellcolor[HTML]{E6F2FF}45.74k & \cellcolor[HTML]{E6F2FF}35.41k & \cellcolor[HTML]{E6F2FF}37.09k & \cellcolor[HTML]{E6F2FF}29.41k & \cellcolor[HTML]{E6F2FF}15.52k & 
\cellcolor[HTML]{E6FFE6}2.46 & \cellcolor[HTML]{E6FFE6}4.92 & \cellcolor[HTML]{E6FFE6}2.57 & \cellcolor[HTML]{E6FFE6}2.68 & \cellcolor[HTML]{E6FFE6}4.45 & 
\cellcolor[HTML]{FFF7E6}19.02 & \cellcolor[HTML]{FFF7E6}3.14 & \cellcolor[HTML]{FFF7E6}17.64 & \cellcolor[HTML]{FFF7E6}17.13 & \cellcolor[HTML]{FFF7E6}7.05 \\
& riscv32i & 
\cellcolor[HTML]{F9F2E7}53.23k & \cellcolor[HTML]{F9F2E7}19.76k & \cellcolor[HTML]{F9F2E7}40.78k & \cellcolor[HTML]{F9F2E7}31.50k & \cellcolor[HTML]{F9F2E7}9.41k & 
\cellcolor[HTML]{E6F2FF}66.91k & \cellcolor[HTML]{E6F2FF}47.74k & \cellcolor[HTML]{E6F2FF}56.38k & \cellcolor[HTML]{E6F2FF}45.79k & \cellcolor[HTML]{E6F2FF}26.04k & 
\cellcolor[HTML]{E6FFE6}2.51 & \cellcolor[HTML]{E6FFE6}4.83 & \cellcolor[HTML]{E6FFE6}2.77 & \cellcolor[HTML]{E6FFE6}2.91 & \cellcolor[HTML]{E6FFE6}5.53 & 
\cellcolor[HTML]{FFF7E6}17.43 & \cellcolor[HTML]{FFF7E6}3.20 & \cellcolor[HTML]{FFF7E6}15.24 & \cellcolor[HTML]{FFF7E6}15.57 & \cellcolor[HTML]{FFF7E6}5.42 \\
& ibex & 
\cellcolor[HTML]{F9F2E7}96.60k & \cellcolor[HTML]{F9F2E7}35.64k & \cellcolor[HTML]{F9F2E7}72.50k & \cellcolor[HTML]{F9F2E7}56.14k & \cellcolor[HTML]{F9F2E7}16.62k & 
\cellcolor[HTML]{E6F2FF}122.18k & \cellcolor[HTML]{E6F2FF}84.69k & \cellcolor[HTML]{E6F2FF}98.88k & \cellcolor[HTML]{E6F2FF}79.63k & \cellcolor[HTML]{E6F2FF}43.94k & 
\cellcolor[HTML]{E6FFE6}2.53 & \cellcolor[HTML]{E6FFE6}4.75 & \cellcolor[HTML]{E6FFE6}2.73 & \cellcolor[HTML]{E6FFE6}2.84 & \cellcolor[HTML]{E6FFE6}5.29 & 
\cellcolor[HTML]{FFF7E6}18.44 & \cellcolor[HTML]{FFF7E6}3.05 & \cellcolor[HTML]{FFF7E6}17.90 & \cellcolor[HTML]{FFF7E6}18.85 & \cellcolor[HTML]{FFF7E6}6.86 \\
& tinyRocket & 
\cellcolor[HTML]{F9F2E7}50.63k & \cellcolor[HTML]{F9F2E7}32.60k & \cellcolor[HTML]{F9F2E7}42.82k & \cellcolor[HTML]{F9F2E7}15.43k & \cellcolor[HTML]{F9F2E7}27.66k & 
\cellcolor[HTML]{E6F2FF}36.27k & \cellcolor[HTML]{E6F2FF}66.69k & \cellcolor[HTML]{E6F2FF}26.03k & \cellcolor[HTML]{E6F2FF}13.54k & \cellcolor[HTML]{E6F2FF}14.90k & 
\cellcolor[HTML]{E6FFE6}1.43 & \cellcolor[HTML]{E6FFE6}4.09 & \cellcolor[HTML]{E6FFE6}1.22 & \cellcolor[HTML]{E6FFE6}1.75 & \cellcolor[HTML]{E6FFE6}1.08 & 
\cellcolor[HTML]{FFF7E6}11.00 & \cellcolor[HTML]{FFF7E6}2.40 & \cellcolor[HTML]{FFF7E6}9.68 & \cellcolor[HTML]{FFF7E6}12.07 & \cellcolor[HTML]{FFF7E6}4.53 \\
& swerv & 
\cellcolor[HTML]{F9F2E7}538.36k & \cellcolor[HTML]{F9F2E7}196.94k & \cellcolor[HTML]{F9F2E7}384.55k & \cellcolor[HTML]{F9F2E7}294.23k & \cellcolor[HTML]{F9F2E7}92.36k & 
\cellcolor[HTML]{E6F2FF}684.89k & \cellcolor[HTML]{E6F2FF}469.98k & \cellcolor[HTML]{E6F2FF}510.70k & \cellcolor[HTML]{E6F2FF}388.58k & \cellcolor[HTML]{E6F2FF}229.49k & 
\cellcolor[HTML]{E6FFE6}2.54 & \cellcolor[HTML]{E6FFE6}4.77 & \cellcolor[HTML]{E6FFE6}2.66 & \cellcolor[HTML]{E6FFE6}2.64 & \cellcolor[HTML]{E6FFE6}4.97 & 
\cellcolor[HTML]{FFF7E6}17.10 & \cellcolor[HTML]{FFF7E6}3.08 & \cellcolor[HTML]{FFF7E6}19.33 & \cellcolor[HTML]{FFF7E6}28.16 & \cellcolor[HTML]{FFF7E6}8.27 \\
& bp\_multi & 
\cellcolor[HTML]{F9F2E7}296.72k & \cellcolor[HTML]{F9F2E7}171.27k & \cellcolor[HTML]{F9F2E7}235.98k & \cellcolor[HTML]{F9F2E7}107.09k & \cellcolor[HTML]{F9F2E7}130.34k & 
\cellcolor[HTML]{E6F2FF}250.15k & \cellcolor[HTML]{E6F2FF}353.77k & \cellcolor[HTML]{E6F2FF}176.42k & \cellcolor[HTML]{E6F2FF}102.51k & \cellcolor[HTML]{E6F2FF}86.96k & 
\cellcolor[HTML]{E6FFE6}1.69 & \cellcolor[HTML]{E6FFE6}4.13 & \cellcolor[HTML]{E6FFE6}1.50 & \cellcolor[HTML]{E6FFE6}1.91 & \cellcolor[HTML]{E6FFE6}1.33 & 
\cellcolor[HTML]{FFF7E6}11.72 & \cellcolor[HTML]{FFF7E6}2.58 & \cellcolor[HTML]{FFF7E6}18.45 & \cellcolor[HTML]{FFF7E6}25.72 & \cellcolor[HTML]{FFF7E6}6.42 \\
\cmidrule{2-22}
 &\textbf{Average} & 
\cellcolor[HTML]{F9F2E7}154.85k & \cellcolor[HTML]{F9F2E7}69.63k & \cellcolor[HTML]{F9F2E7}118.58k & \cellcolor[HTML]{F9F2E7}78.52k & \cellcolor[HTML]{F9F2E7}29.88k & 
\cellcolor[HTML]{E6F2FF}177.80k & \cellcolor[HTML]{E6F2FF}155.03k & \cellcolor[HTML]{E6F2FF}163.18k & \cellcolor[HTML]{E6F2FF}114.28k & \cellcolor[HTML]{E6F2FF}61.65k & 
\cellcolor[HTML]{E6FFE6}2.15 & \cellcolor[HTML]{E6FFE6}4.51 & \cellcolor[HTML]{E6FFE6}2.17 & \cellcolor[HTML]{E6FFE6}2.41 & \cellcolor[HTML]{E6FFE6}3.90 & 
\cellcolor[HTML]{FFF7E6}15.90 & \cellcolor[HTML]{FFF7E6}2.94 & \cellcolor[HTML]{FFF7E6}16.10 & \cellcolor[HTML]{FFF7E6}18.64 & \cellcolor[HTML]{FFF7E6}6.36 \\
\midrule

\multirow{12}{*}{\rotatebox[origin=c]{90}{\textbf{Comm. ctrl.}}} 
& uart & 
\cellcolor[HTML]{F9F2E7}2.73k & \cellcolor[HTML]{F9F2E7}1.11k & \cellcolor[HTML]{F9F2E7}1.94k & \cellcolor[HTML]{F9F2E7}1.48k & \cellcolor[HTML]{F9F2E7}0.52k & 
\cellcolor[HTML]{E6F2FF}3.29k & \cellcolor[HTML]{E6F2FF}2.30k & \cellcolor[HTML]{E6F2FF}2.60k & \cellcolor[HTML]{E6F2FF}1.88k & \cellcolor[HTML]{E6F2FF}1.15k & 
\cellcolor[HTML]{E6FFE6}2.41 & \cellcolor[HTML]{E6FFE6}4.13 & \cellcolor[HTML]{E6FFE6}2.67 & \cellcolor[HTML]{E6FFE6}2.54 & \cellcolor[HTML]{E6FFE6}4.47 & 
\cellcolor[HTML]{FFF7E6}13.80 & \cellcolor[HTML]{FFF7E6}3.27 & \cellcolor[HTML]{FFF7E6}14.79 & \cellcolor[HTML]{FFF7E6}10.61 & \cellcolor[HTML]{FFF7E6}5.63 \\
& sasc\_top & 
\cellcolor[HTML]{F9F2E7}3.45k & \cellcolor[HTML]{F9F2E7}1.38k & \cellcolor[HTML]{F9F2E7}2.44k & \cellcolor[HTML]{F9F2E7}1.84k & \cellcolor[HTML]{F9F2E7}0.63k & 
\cellcolor[HTML]{E6F2FF}4.17k & \cellcolor[HTML]{E6F2FF}2.88k & \cellcolor[HTML]{E6F2FF}3.01k & \cellcolor[HTML]{E6F2FF}2.12k & \cellcolor[HTML]{E6F2FF}1.33k & 
\cellcolor[HTML]{E6FFE6}2.42 & \cellcolor[HTML]{E6FFE6}4.18 & \cellcolor[HTML]{E6FFE6}2.47 & \cellcolor[HTML]{E6FFE6}2.31 & \cellcolor[HTML]{E6FFE6}4.22 & 
\cellcolor[HTML]{FFF7E6}12.20 & \cellcolor[HTML]{FFF7E6}2.88 & \cellcolor[HTML]{FFF7E6}12.69 & \cellcolor[HTML]{FFF7E6}15.79 & \cellcolor[HTML]{FFF7E6}4.75 \\
& i2c\_verilog & 
\cellcolor[HTML]{F9F2E7}4.69k & \cellcolor[HTML]{F9F2E7}1.85k & \cellcolor[HTML]{F9F2E7}3.39k & \cellcolor[HTML]{F9F2E7}2.56k & \cellcolor[HTML]{F9F2E7}0.87k & 
\cellcolor[HTML]{E6F2FF}5.70k & \cellcolor[HTML]{E6F2FF}4.03k & \cellcolor[HTML]{E6F2FF}6.54k & \cellcolor[HTML]{E6F2FF}5.28k & \cellcolor[HTML]{E6F2FF}4.12k & 
\cellcolor[HTML]{E6FFE6}2.43 & \cellcolor[HTML]{E6FFE6}4.35 & \cellcolor[HTML]{E6FFE6}3.86 & \cellcolor[HTML]{E6FFE6}4.13 & \cellcolor[HTML]{E6FFE6}9.51 & 
\cellcolor[HTML]{FFF7E6}13.42 & \cellcolor[HTML]{FFF7E6}3.02 & \cellcolor[HTML]{FFF7E6}11.96 & \cellcolor[HTML]{FFF7E6}13.93 & \cellcolor[HTML]{FFF7E6}5.49 \\
& simple\_spi\_top & 
\cellcolor[HTML]{F9F2E7}4.57k & \cellcolor[HTML]{F9F2E7}1.84k & \cellcolor[HTML]{F9F2E7}3.35k & \cellcolor[HTML]{F9F2E7}2.52k & \cellcolor[HTML]{F9F2E7}0.85k & 
\cellcolor[HTML]{E6F2FF}5.49k & \cellcolor[HTML]{E6F2FF}3.89k & \cellcolor[HTML]{E6F2FF}4.17k & \cellcolor[HTML]{E6F2FF}3.01k & \cellcolor[HTML]{E6F2FF}1.80k & 
\cellcolor[HTML]{E6FFE6}2.40 & \cellcolor[HTML]{E6FFE6}4.23 & \cellcolor[HTML]{E6FFE6}2.49 & \cellcolor[HTML]{E6FFE6}2.39 & \cellcolor[HTML]{E6FFE6}4.23 & 
\cellcolor[HTML]{FFF7E6}13.33 & \cellcolor[HTML]{FFF7E6}2.90 & \cellcolor[HTML]{FFF7E6}12.23 & \cellcolor[HTML]{FFF7E6}13.55 & \cellcolor[HTML]{FFF7E6}4.93 \\
& spi\_top & 
\cellcolor[HTML]{F9F2E7}15.62k & \cellcolor[HTML]{F9F2E7}6.07k & \cellcolor[HTML]{F9F2E7}12.31k & \cellcolor[HTML]{F9F2E7}9.47k & \cellcolor[HTML]{F9F2E7}2.93k & 
\cellcolor[HTML]{E6F2FF}19.18k & \cellcolor[HTML]{E6F2FF}14.21k & \cellcolor[HTML]{E6F2FF}19.15k & \cellcolor[HTML]{E6F2FF}15.39k & \cellcolor[HTML]{E6F2FF}9.99k & 
\cellcolor[HTML]{E6FFE6}2.46 & \cellcolor[HTML]{E6FFE6}4.68 & \cellcolor[HTML]{E6FFE6}3.11 & \cellcolor[HTML]{E6FFE6}3.25 & \cellcolor[HTML]{E6FFE6}6.81 & 
\cellcolor[HTML]{FFF7E6}16.79 & \cellcolor[HTML]{FFF7E6}3.02 & \cellcolor[HTML]{FFF7E6}13.75 & \cellcolor[HTML]{FFF7E6}14.99 & \cellcolor[HTML]{FFF7E6}5.68 \\
& dynamic\_node & 
\cellcolor[HTML]{F9F2E7}67.87k & \cellcolor[HTML]{F9F2E7}26.28k & \cellcolor[HTML]{F9F2E7}46.91k & \cellcolor[HTML]{F9F2E7}35.50k & \cellcolor[HTML]{F9F2E7}12.10k & 
\cellcolor[HTML]{E6F2FF}83.87k & \cellcolor[HTML]{E6F2FF}56.29k & \cellcolor[HTML]{E6F2FF}57.47k & \cellcolor[HTML]{E6F2FF}40.55k & \cellcolor[HTML]{E6F2FF}24.88k & 
\cellcolor[HTML]{E6FFE6}2.47 & \cellcolor[HTML]{E6FFE6}4.28 & \cellcolor[HTML]{E6FFE6}2.45 & \cellcolor[HTML]{E6FFE6}2.28 & \cellcolor[HTML]{E6FFE6}4.11 & 
\cellcolor[HTML]{FFF7E6}15.10 & \cellcolor[HTML]{FFF7E6}3.22 & \cellcolor[HTML]{FFF7E6}16.09 & \cellcolor[HTML]{FFF7E6}15.00 & \cellcolor[HTML]{FFF7E6}6.62 \\
& pci & 
\cellcolor[HTML]{F9F2E7}72.69k & \cellcolor[HTML]{F9F2E7}27.53k & \cellcolor[HTML]{F9F2E7}48.45k & \cellcolor[HTML]{F9F2E7}36.59k & \cellcolor[HTML]{F9F2E7}12.23k & 
\cellcolor[HTML]{E6F2FF}90.70k & \cellcolor[HTML]{E6F2FF}56.01k & \cellcolor[HTML]{E6F2FF}72.86k & \cellcolor[HTML]{E6F2FF}51.99k & \cellcolor[HTML]{E6F2FF}38.15k & 
\cellcolor[HTML]{E6FFE6}2.50 & \cellcolor[HTML]{E6FFE6}4.07 & \cellcolor[HTML]{E6FFE6}3.01 & \cellcolor[HTML]{E6FFE6}2.84 & \cellcolor[HTML]{E6FFE6}6.24 & 
\cellcolor[HTML]{FFF7E6}15.13 & \cellcolor[HTML]{FFF7E6}2.88 & \cellcolor[HTML]{FFF7E6}16.35 & \cellcolor[HTML]{FFF7E6}25.42 & \cellcolor[HTML]{FFF7E6}6.37 \\
& usb\_funct & 
\cellcolor[HTML]{F9F2E7}67.30k & \cellcolor[HTML]{F9F2E7}26.63k & \cellcolor[HTML]{F9F2E7}48.60k & \cellcolor[HTML]{F9F2E7}36.55k & \cellcolor[HTML]{F9F2E7}12.30k & 
\cellcolor[HTML]{E6F2FF}81.58k & \cellcolor[HTML]{E6F2FF}58.85k & \cellcolor[HTML]{E6F2FF}60.03k & \cellcolor[HTML]{E6F2FF}44.10k & \cellcolor[HTML]{E6F2FF}23.78k & 
\cellcolor[HTML]{E6FFE6}2.42 & \cellcolor[HTML]{E6FFE6}4.42 & \cellcolor[HTML]{E6FFE6}2.47 & \cellcolor[HTML]{E6FFE6}2.41 & \cellcolor[HTML]{E6FFE6}3.87 & 
\cellcolor[HTML]{FFF7E6}21.41 & \cellcolor[HTML]{FFF7E6}3.33 & \cellcolor[HTML]{FFF7E6}19.13 & \cellcolor[HTML]{FFF7E6}21.79 & \cellcolor[HTML]{FFF7E6}7.43 \\
& wb\_dma\_top & 
\cellcolor[HTML]{F9F2E7}19.52k & \cellcolor[HTML]{F9F2E7}8.22k & \cellcolor[HTML]{F9F2E7}14.60k & \cellcolor[HTML]{F9F2E7}11.08k & \cellcolor[HTML]{F9F2E7}3.95k & 
\cellcolor[HTML]{E6F2FF}23.03k & \cellcolor[HTML]{E6F2FF}16.72k & \cellcolor[HTML]{E6F2FF}17.79k & \cellcolor[HTML]{E6F2FF}12.52k & \cellcolor[HTML]{E6F2FF}7.62k & 
\cellcolor[HTML]{E6FFE6}2.36 & \cellcolor[HTML]{E6FFE6}4.07 & \cellcolor[HTML]{E6FFE6}2.44 & \cellcolor[HTML]{E6FFE6}2.26 & \cellcolor[HTML]{E6FFE6}3.86 & 
\cellcolor[HTML]{FFF7E6}15.38 & \cellcolor[HTML]{FFF7E6}3.17 & \cellcolor[HTML]{FFF7E6}13.09 & \cellcolor[HTML]{FFF7E6}15.66 & \cellcolor[HTML]{FFF7E6}5.76 \\
& wb\_conmax & 
\cellcolor[HTML]{F9F2E7}181.45k & \cellcolor[HTML]{F9F2E7}71.84k & \cellcolor[HTML]{F9F2E7}144.40k & \cellcolor[HTML]{F9F2E7}110.71k & \cellcolor[HTML]{F9F2E7}36.24k & 
\cellcolor[HTML]{E6F2FF}221.76k & \cellcolor[HTML]{E6F2FF}177.04k & \cellcolor[HTML]{E6F2FF}183.96k & \cellcolor[HTML]{E6F2FF}145.80k & \cellcolor[HTML]{E6F2FF}76.57k & 
\cellcolor[HTML]{E6FFE6}2.44 & \cellcolor[HTML]{E6FFE6}4.93 & \cellcolor[HTML]{E6FFE6}2.55 & \cellcolor[HTML]{E6FFE6}2.63 & \cellcolor[HTML]{E6FFE6}4.23 & 
\cellcolor[HTML]{FFF7E6}24.35 & \cellcolor[HTML]{FFF7E6}2.96 & \cellcolor[HTML]{FFF7E6}21.46 & \cellcolor[HTML]{FFF7E6}18.56 & \cellcolor[HTML]{FFF7E6}9.04 \\
& usb\_phy & 
\cellcolor[HTML]{F9F2E7}2.73k & \cellcolor[HTML]{F9F2E7}1.13k & \cellcolor[HTML]{F9F2E7}1.84k & \cellcolor[HTML]{F9F2E7}1.36k & \cellcolor[HTML]{F9F2E7}0.51k & 
\cellcolor[HTML]{E6F2FF}3.22k & \cellcolor[HTML]{E6F2FF}2.35k & \cellcolor[HTML]{E6F2FF}2.14k & \cellcolor[HTML]{E6F2FF}1.42k & \cellcolor[HTML]{E6F2FF}0.91k & 
\cellcolor[HTML]{E6FFE6}2.36 & \cellcolor[HTML]{E6FFE6}4.15 & \cellcolor[HTML]{E6FFE6}2.33 & \cellcolor[HTML]{E6FFE6}2.09 & \cellcolor[HTML]{E6FFE6}3.56 & 
\cellcolor[HTML]{FFF7E6}10.67 & \cellcolor[HTML]{FFF7E6}3.27 & \cellcolor[HTML]{FFF7E6}11.10 & \cellcolor[HTML]{FFF7E6}11.89 & \cellcolor[HTML]{FFF7E6}4.41 \\
\cmidrule{2-22}
 &\textbf{Average} & 
\cellcolor[HTML]{F9F2E7}42.20k & \cellcolor[HTML]{F9F2E7}16.68k & \cellcolor[HTML]{F9F2E7}31.93k & \cellcolor[HTML]{F9F2E7}24.40k & \cellcolor[HTML]{F9F2E7}8.76k & 
\cellcolor[HTML]{E6F2FF}50.89k & \cellcolor[HTML]{E6F2FF}39.33k & \cellcolor[HTML]{E6F2FF}43.32k & \cellcolor[HTML]{E6F2FF}33.03k & \cellcolor[HTML]{E6F2FF}19.54k & 
\cellcolor[HTML]{E6FFE6}2.42 & \cellcolor[HTML]{E6FFE6}4.30 & \cellcolor[HTML]{E6FFE6}2.67 & \cellcolor[HTML]{E6FFE6}2.61 & \cellcolor[HTML]{E6FFE6}4.78 & 
\cellcolor[HTML]{FFF7E6}15.65 & \cellcolor[HTML]{FFF7E6}3.06 & \cellcolor[HTML]{FFF7E6}14.71 & \cellcolor[HTML]{FFF7E6}16.23 & \cellcolor[HTML]{FFF7E6}5.98 \\
\midrule

\multirow{4}{*}{\rotatebox[origin=c]{90}{\textbf{DSP Core}}} 
& fir & 
\cellcolor[HTML]{F9F2E7}22.63k & \cellcolor[HTML]{F9F2E7}8.82k & \cellcolor[HTML]{F9F2E7}16.90k & \cellcolor[HTML]{F9F2E7}12.98k & \cellcolor[HTML]{F9F2E7}3.98k & 
\cellcolor[HTML]{E6F2FF}27.68k & \cellcolor[HTML]{E6F2FF}19.51k & \cellcolor[HTML]{E6F2FF}21.46k & \cellcolor[HTML]{E6F2FF}16.28k & \cellcolor[HTML]{E6F2FF}8.80k & 
\cellcolor[HTML]{E6FFE6}2.45 & \cellcolor[HTML]{E6FFE6}4.42 & \cellcolor[HTML]{E6FFE6}2.54 & \cellcolor[HTML]{E6FFE6}2.51 & \cellcolor[HTML]{E6FFE6}4.42 & 
\cellcolor[HTML]{FFF7E6}17.94 & \cellcolor[HTML]{FFF7E6}3.07 & \cellcolor[HTML]{FFF7E6}16.16 & \cellcolor[HTML]{FFF7E6}20.45 & \cellcolor[HTML]{FFF7E6}6.23 \\
& jpeg & 
\cellcolor[HTML]{F9F2E7}331.18k & \cellcolor[HTML]{F9F2E7}134.01k & \cellcolor[HTML]{F9F2E7}244.06k & \cellcolor[HTML]{F9F2E7}183.26k & \cellcolor[HTML]{F9F2E7}60.85k & 
\cellcolor[HTML]{E6F2FF}394.38k & \cellcolor[HTML]{E6F2FF}301.36k & \cellcolor[HTML]{E6F2FF}297.27k & \cellcolor[HTML]{E6F2FF}236.65k & \cellcolor[HTML]{E6F2FF}118.01k & 
\cellcolor[HTML]{E6FFE6}2.38 & \cellcolor[HTML]{E6FFE6}4.50 & \cellcolor[HTML]{E6FFE6}2.44 & \cellcolor[HTML]{E6FFE6}2.58 & \cellcolor[HTML]{E6FFE6}3.88 & 
\cellcolor[HTML]{FFF7E6}23.11 & \cellcolor[HTML]{FFF7E6}3.04 & \cellcolor[HTML]{FFF7E6}23.11 & \cellcolor[HTML]{FFF7E6}24.50 & \cellcolor[HTML]{FFF7E6}8.91 \\
& idft & 
\cellcolor[HTML]{F9F2E7}817.88k & \cellcolor[HTML]{F9F2E7}313.24k & \cellcolor[HTML]{F9F2E7}527.42k & \cellcolor[HTML]{F9F2E7}392.61k & \cellcolor[HTML]{F9F2E7}134.94k & 
\cellcolor[HTML]{E6F2FF}1009.40k & \cellcolor[HTML]{E6F2FF}622.01k & \cellcolor[HTML]{E6F2FF}590.78k & \cellcolor[HTML]{E6F2FF}382.70k & \cellcolor[HTML]{E6F2FF}236.06k & 
\cellcolor[HTML]{E6FFE6}2.47 & \cellcolor[HTML]{E6FFE6}3.97 & \cellcolor[HTML]{E6FFE6}2.24 & \cellcolor[HTML]{E6FFE6}1.91 & \cellcolor[HTML]{E6FFE6}3.50 & 
\cellcolor[HTML]{FFF7E6}13.06 & \cellcolor[HTML]{FFF7E6}3.26 & \cellcolor[HTML]{FFF7E6}19.57 & \cellcolor[HTML]{FFF7E6}26.46 & \cellcolor[HTML]{FFF7E6}6.64 \\
\cmidrule{2-22}
 &\textbf{Average} & 
\cellcolor[HTML]{F9F2E7}390.56k & \cellcolor[HTML]{F9F2E7}152.02k & \cellcolor[HTML]{F9F2E7}262.79k & \cellcolor[HTML]{F9F2E7}196.28k & \cellcolor[HTML]{F9F2E7}66.59k & 
\cellcolor[HTML]{E6F2FF}477.15k & \cellcolor[HTML]{E6F2FF}314.29k & \cellcolor[HTML]{E6F2FF}303.17k & \cellcolor[HTML]{E6F2FF}211.88k & \cellcolor[HTML]{E6F2FF}120.96k & 
\cellcolor[HTML]{E6FFE6}2.43 & \cellcolor[HTML]{E6FFE6}4.30 & \cellcolor[HTML]{E6FFE6}2.41 & \cellcolor[HTML]{E6FFE6}2.33 & \cellcolor[HTML]{E6FFE6}3.93 & 
\cellcolor[HTML]{FFF7E6}18.04 & \cellcolor[HTML]{FFF7E6}3.12 & \cellcolor[HTML]{FFF7E6}19.61 & \cellcolor[HTML]{FFF7E6}23.80 & \cellcolor[HTML]{FFF7E6}7.26 \\
\midrule

\multicolumn{2}{c|}{\textbf{Total Average}} & 
\cellcolor[HTML]{F9F2E7}124.41k & \cellcolor[HTML]{F9F2E7}49.92k & \cellcolor[HTML]{F9F2E7}89.64k & \cellcolor[HTML]{F9F2E7}64.98k & \cellcolor[HTML]{F9F2E7}25.06k & 
\cellcolor[HTML]{E6F2FF}149.29k & \cellcolor[HTML]{E6F2FF}110.17k & \cellcolor[HTML]{E6F2FF}107.69k & \cellcolor[HTML]{E6F2FF}78.79k & \cellcolor[HTML]{E6F2FF}46.60k & 
\cellcolor[HTML]{E6FFE6}2.40 & \cellcolor[HTML]{E6FFE6}4.48 & \cellcolor[HTML]{E6FFE6}2.55 & \cellcolor[HTML]{E6FFE6}2.56 & \cellcolor[HTML]{E6FFE6}4.58 & 
\cellcolor[HTML]{FFF7E6}16.43 & \cellcolor[HTML]{FFF7E6}3.04 & \cellcolor[HTML]{FFF7E6}16.16 & \cellcolor[HTML]{FFF7E6}18.25 & \cellcolor[HTML]{FFF7E6}6.35 \\
\bottomrule
\end{tabular}%
}
\end{table*}

This appendix summarizes structural statistics for 30 circuits across five graph views (b--f): $\#\mathrm{nodes}$, $\#\mathrm{edges}$, $\mathrm{avg\_degree}$, and $\mathrm{avg\_shortest\ path}$ computed by a unified late-stage pipeline. \autoref{tab:view_stats} shows consistent patterns aligned with physical design: view~(b) has \emph{moderate} degree ($\approx$2.4--2.6) and \emph{long} paths ($\approx$11--23); view~(c) is the \emph{densest} node-centric option (higher degree, very short paths); views~(d)/(e) retain \emph{moderate} degrees with \emph{mid-to-long}/\emph{very long} diameters; view~(f) is \emph{dense, small-world} (high degree, short paths). These structures match trends in \autoref{tab:placement_views} and \autoref{tab:routing_results}.

\subsection{Why view~(b) leads and node-centric views prevail}
\subsubsection{Average Statistics}
view~(b) balances locality and reach: \emph{moderate degree} (\(\approx\)2.40) with \emph{long paths} (\(\approx\)16.43). A moderate degree constrains neighborhood breadth—reducing noise amplification and over-mixing—while long paths preserve global dependencies needed for macro-to-cell interactions and net connectivity. view~(c) is the \emph{densest} node-centric option (degree \(\approx\)4.48, paths \(\approx\)3.04): high degree and very short paths concentrate information locally; this aids fine-grained aggregation but compresses structural distance, limiting long-range context when global coordination is required. views~(d)/(e) retain \emph{moderate degrees} (\(\approx\)2.55/2.56) and \emph{mid-to-long paths} (\(\approx\)16.16/18.25), reflecting pin-level dependencies along nets: multi-hop propagation is supported without saturating neighborhoods. view~(f) is \emph{dense, small-world} (degree \(\approx\)4.58, paths \(\approx\)6.35), tying distant subcircuits through hubs; this increases correlation but can blur constraint boundaries and inject hub bias.

\subsubsection{Cross-Stage Outcomes}
Placement favors node-centric graphs: view~(b) is strongest, with view~(c) close. Placement decisions rely on global context and net topology; B’s long paths and controlled degree deliver broad context without noise, explaining its margin. C competes when local density helps, but its short-path bias can hamper cases needing global coordination. In routing, view~(b) remains strongest overall, while among edge-centric views, views~(d)/(e) are competitive: their mid-to-long paths track channel capacity and congestion cascades along connections, making them useful complements for routing-oriented analysis. view~(f) underperforms because hub-dominated mixing erodes locality and misaligns with edge constraints.

\subsubsection{Summaries}
Prefer view~(b) as a robust default across stages. Use view~(c) judiciously for workloads dominated by local interactions; treat views~(d)/(e) as complementary choices when routing fidelity is prioritized. Avoid view~(f) for headline reporting due to small-world mixing and hub bias that harm both placement coordination and routing locality.

\subsection{Why view~(d) leads among edge-centric views: statistical comparison}\label{sec:supp_view_d_edge}

\subsubsection{Key Statistics}
view~(d) combines \emph{moderate degree} with \emph{mid-to-long paths}, yielding expressive yet unsaturated neighborhoods. Compared with view~(e), degrees are similar on average, but E’s paths are consistently longer, shifting useful interactions to larger hop distances. view~(f) exhibits \emph{higher degrees} and \emph{shorter paths}, encouraging small-world mixing and hub bias that blur fine-grained signals. In edges-to-nodes balance, view~(d) remains near-linear, whereas view~(f) increases density and alters neighborhood selectivity.

\subsubsection{Cross-Stage Alignment}
For routing-oriented analysis, view~(d) is the strongest edge-centric complement; view~(e) supports longer-range effects when they are critical; view~(f)’s small-world structure is generally misaligned.

\subsubsection{Summaries}
Average structural statistics explain stage-level outcomes without model-specific details: view~(b) is robust across placement and routing; view~(c) is strong but less stable; views~(d)/(e) are viable edge-centric complements; view~(f) is not recommended for headline results.

\section{Hyperparameter Tuning}
\label{sec:hp_tune}

The experimental results for the Placement Task across different graph views (b-f), as presented in \autoref{tab:placement_views} of the main paper, and for the Routing Task in \autoref{tab:routing_results}, were obtained using the specific hyperparameter configurations detailed in \autoref{tab:placement_hparams} and \autoref{tab:routing_hparams}, respectively. To ensure a fair and consistent comparison across the different graph views, each model (GINE, ResGatedGCN, GAT) utilized a single, fixed set of hyperparameters for all views within a given task. This means that the same hyperparameter values were applied to all views (b), (c), (d), (e), and (f) for the Placement task, and similarly for the Routing task, without further view-specific tuning.

\begin{table}[tbp]
\centering
\caption{Placement Task Hyperparameters for Different Graph Views}
\label{tab:placement_hparams}
\resizebox{\linewidth}{!}{%
\begin{tabular}{lcccccccc}
\toprule
\textbf{Hyper} & \textbf{lr} & \textbf{wd} & \textbf{hid\_dim} & \textbf{act} & \textbf{gnn\_layer} & \textbf{dropout} & \textbf{head\_layer} & \textbf{head\_dim} \\
\midrule

\multicolumn{9}{c}{\cellcolor[HTML]{E6F2FF}\textbf{(b)}} \\
\cmidrule{1-9}
\rowcolor[HTML]{FFF7E6} GINE & 1e-4 & 1e-4 & 256 & relu & 4 & 0.3 & 2 & 256 \\
\rowcolor[HTML]{FFF7E6} ResGatedGCN & 1e-4 & 1e-4 & 256 & relu & 4 & 0.3 & 2 & 256 \\
\rowcolor[HTML]{FFF7E6} GAT & 1e-4 & 1e-4 & 256 & relu & 4 & 0.3 & 2 & 256 \\

\multicolumn{9}{c}{\cellcolor[HTML]{E6F2FF}\textbf{(c)}} \\
\cmidrule{1-9}
\rowcolor[HTML]{FFF7E6} GINE & 1e-4 & 1e-4 & 256 & relu & 4 & 0.3 & 2 & 256 \\
\rowcolor[HTML]{FFF7E6} ResGatedGCN & 1e-4 & 1e-4 & 256 & relu & 4 & 0.3 & 2 & 256 \\
\rowcolor[HTML]{FFF7E6} GAT & 1e-4 & 1e-4 & 256 & relu & 4 & 0.3 & 2 & 256 \\

\multicolumn{9}{c}{\cellcolor[HTML]{E6F2FF}\textbf{(d)}} \\
\cmidrule{1-9}
\rowcolor[HTML]{FFF7E6} GINE & 5e-4 & 1e-5 & 256 & leakyrelu & 4 & 0.3 & 2 & 256 \\
\rowcolor[HTML]{FFF7E6} ResGatedGCN & 5e-4 & 1e-5 & 256 & leakyrelu & 4 & 0.3 & 2 & 256 \\
\rowcolor[HTML]{FFF7E6} GAT & 5e-4 & 1e-5 & 256 & leakyrelu & 4 & 0.3 & 2 & 256 \\

\multicolumn{9}{c}{\cellcolor[HTML]{E6F2FF}\textbf{(e)}} \\
\cmidrule{1-9}
\rowcolor[HTML]{FFF7E6} GINE & 5e-4 & 1e-5 & 256 & leakyrelu & 4 & 0.3 & 2 & 256 \\
\rowcolor[HTML]{FFF7E6} ResGatedGCN & 5e-4 & 1e-5 & 256 & leakyrelu & 4 & 0.3 & 2 & 256 \\
\rowcolor[HTML]{FFF7E6} GAT & 5e-4 & 1e-5 & 256 & leakyrelu & 4 & 0.3 & 2 & 256 \\

\multicolumn{9}{c}{\cellcolor[HTML]{E6F2FF}\textbf{(f)}} \\
\cmidrule{1-9}
\rowcolor[HTML]{FFF7E6} GINE & 5e-4 & 1e-5 & 256 & leakyrelu & 4 & 0.3 & 2 & 256 \\
\rowcolor[HTML]{FFF7E6} ResGatedGCN & 5e-4 & 1e-5 & 256 & leakyrelu & 4 & 0.3 & 2 & 256 \\
\rowcolor[HTML]{FFF7E6} GAT & 5e-4 & 1e-5 & 256 & leakyrelu & 4 & 0.3 & 2 & 256 \\
\bottomrule
\end{tabular}%
}
\end{table}

\begin{table}[tbp]
\centering
\caption{Routing Task Hyperparameters for Different Graph Views}
\label{tab:routing_hparams}
\resizebox{\linewidth}{!}{%
\begin{tabular}{lcccccccc}
\toprule
\textbf{Hyper} & \textbf{lr} & \textbf{wd} & \textbf{hid\_dim} & \textbf{act} & \textbf{gnn\_layer} & \textbf{dropout} & \textbf{head\_layer} & \textbf{head\_dim} \\
\midrule

\multicolumn{9}{c}{\cellcolor[HTML]{E6F2FF}\textbf{(b)}} \\
\cmidrule{1-9}
\rowcolor[HTML]{FFF7E6} GINE & 1e-4 & 1e-4 & 256 & leakyrelu & 4 & 0.3 & 2 & 256 \\
\rowcolor[HTML]{FFF7E6} ResGatedGCN & 1e-4 & 1e-4 & 256 & leakyrelu & 4 & 0.3 & 2 & 256 \\
\rowcolor[HTML]{FFF7E6} GAT & 1e-4 & 1e-4 & 256 & leakyrelu & 4 & 0.3 & 2 & 256 \\

\multicolumn{9}{c}{\cellcolor[HTML]{E6F2FF}\textbf{(c)}} \\
\cmidrule{1-9}
\rowcolor[HTML]{FFF7E6} GINE & 1e-4 & 1e-4 & 256 & leakyrelu & 4 & 0.3 & 2 & 256 \\
\rowcolor[HTML]{FFF7E6} ResGatedGCN & 1e-4 & 1e-4 & 256 & leakyrelu & 4 & 0.3 & 2 & 256 \\
\rowcolor[HTML]{FFF7E6} GAT & 1e-4 & 1e-4 & 256 & leakyrelu & 4 & 0.3 & 2 & 256 \\

\multicolumn{9}{c}{\cellcolor[HTML]{E6F2FF}\textbf{(d)}} \\
\cmidrule{1-9}
\rowcolor[HTML]{FFF7E6} GINE & 5e-4 & 1e-5 & 256 & leakyrelu & 4 & 0.3 & 2 & 256 \\
\rowcolor[HTML]{FFF7E6} ResGatedGCN & 5e-4 & 1e-5 & 256 & leakyrelu & 4 & 0.3 & 2 & 256 \\
\rowcolor[HTML]{FFF7E6} GAT & 5e-4 & 1e-5 & 256 & leakyrelu & 4 & 0.3 & 2 & 256 \\

\multicolumn{9}{c}{\cellcolor[HTML]{E6F2FF}\textbf{(e)}} \\
\cmidrule{1-9}
\rowcolor[HTML]{FFF7E6} GINE & 5e-4 & 1e-5 & 256 & leakyrelu & 4 & 0.3 & 2 & 256 \\
\rowcolor[HTML]{FFF7E6} ResGatedGCN & 5e-4 & 1e-5 & 256 & leakyrelu & 4 & 0.3 & 2 & 256 \\
\rowcolor[HTML]{FFF7E6} GAT & 5e-4 & 1e-5 & 256 & leakyrelu & 4 & 0.3 & 2 & 256 \\

\multicolumn{9}{c}{\cellcolor[HTML]{E6F2FF}\textbf{(f)}} \\
\cmidrule{1-9}
\rowcolor[HTML]{FFF7E6} GINE & 5e-4 & 1e-5 & 256 & leakyrelu & 4 & 0.3 & 2 & 256 \\
\rowcolor[HTML]{FFF7E6} ResGatedGCN & 5e-4 & 1e-5 & 256 & leakyrelu & 4 & 0.3 & 2 & 256 \\
\rowcolor[HTML]{FFF7E6} GAT & 5e-4 & 1e-5 & 256 & leakyrelu & 4 & 0.3 & 2 & 256 \\
\bottomrule
\end{tabular}%
}
\end{table}

\section{GNN Depth and Head Depth}
\label{sec:gnn_head}
\begin{table}[h!]         
  \centering
  \caption{Placement on view~(d) vs.\ GNN depth (3--6).}
  \label{tab:place_d_gnnlayers}
  \resizebox{\linewidth}{!}{%
\begin{tabular}{c| 
    >{\columncolor[HTML]{FFF7E6}}c >{\columncolor[HTML]{FFF7E6}}c >{\columncolor[HTML]{FFF7E6}}c >{\columncolor[HTML]{FFF7E6}}c 
    >{\columncolor[HTML]{E6F2FF}}c >{\columncolor[HTML]{E6F2FF}}c >{\columncolor[HTML]{E6F2FF}}c >{\columncolor[HTML]{E6F2FF}}c 
    }
\toprule
& \multicolumn{4}{c}{\cellcolor[HTML]{FFF7E6}\textbf{GINE}} & \multicolumn{4}{c}{\cellcolor[HTML]{E6F2FF}\textbf{ResGatedGCN}} \\
    \cmidrule(lr){2-5} \cmidrule(lr){6-9}
    \textbf{\#layers} & \textbf{MAE$\downarrow$} & \textbf{RMSE$\downarrow$} & \textbf{R$^2\uparrow$} & \textbf{\#Param.} &
    \textbf{MAE$\downarrow$} & \textbf{RMSE$\downarrow$} & \textbf{R$^2\uparrow$} & \textbf{\#Param.} \\
    \midrule
    \textbf{3} & 0.4552 & 0.6098 & 0.6811 & 0.83M & \textbf{0.4521} & \textbf{0.6026} & \textbf{0.6886} & 1.62M \\
    \textbf{4} & 0.4451 & 0.5937 & 0.6975 & 1.03M & 0.4670 & 0.6197 & 0.6704 & 2.08M \\
    \textbf{5} & 0.4837 & 0.6467 & 0.6414 & 1.23M & 0.4700 & 0.6077 & 0.6833 & 2.54M \\
    \textbf{6} & \textbf{0.4063} & \textbf{0.5478} & \textbf{0.7427} & 1.42M & 0.5449 & 0.6786 & 0.6051 & 3.00M \\
    \bottomrule
  \end{tabular}%
  }
\end{table}

\begin{table}[h!]         
  \centering
  \caption{Routing on view~(d) vs.\ GNN depth (3--6).}
  \label{tab:route_d_gnnlayers_test}
  \resizebox{\linewidth}{!}{%
\begin{tabular}{c| 
    >{\columncolor[HTML]{FFF7E6}}c >{\columncolor[HTML]{FFF7E6}}c >{\columncolor[HTML]{FFF7E6}}c >{\columncolor[HTML]{FFF7E6}}c 
    >{\columncolor[HTML]{E6F2FF}}c >{\columncolor[HTML]{E6F2FF}}c >{\columncolor[HTML]{E6F2FF}}c >{\columncolor[HTML]{E6F2FF}}c 
    }
\toprule
& \multicolumn{4}{c}{\cellcolor[HTML]{FFF7E6}\textbf{GINE}} & \multicolumn{4}{c}{\cellcolor[HTML]{E6F2FF}\textbf{ResGatedGCN}} \\
    \cmidrule(lr){2-5} \cmidrule(lr){6-9}
    \textbf{\#layers} & \textbf{MAE$\downarrow$} & \textbf{RMSE$\downarrow$} & \textbf{R$^2\uparrow$} & \textbf{\#Param.} &
    \textbf{MAE$\downarrow$} & \textbf{RMSE$\downarrow$} & \textbf{R$^2\uparrow$} & \textbf{\#Param.} \\
    \midrule
    \textbf{3} & 0.5470 & 0.7172 & 0.6000 & 0.83M & 0.5732 & 0.7557 & 0.5558 & 1.62M \\
    \textbf{4} & 0.6111 & 0.7920 & 0.5122 & 1.03M & 0.6007 & 0.7826 & 0.5237 & 2.08M \\
    \textbf{5} & \textbf{0.5062} & \textbf{0.6805} & \textbf{0.6398} & 1.22M & \textbf{0.5413} & \textbf{0.7091} & \textbf{0.6090} & 2.54M \\
    \textbf{6} & 0.5964 & 0.7571 & 0.5542 & 1.42M & 0.6803 & 0.8942 & 0.3782 & 3.00M \\
    \bottomrule
  \end{tabular}%
  }
\end{table}

\begin{table}[h!]         
  \centering
  \caption{Placement on view~(d) vs.\ head layers (1--4).}
  \label{tab:place_d_headlayers_test}
  \resizebox{\linewidth}{!}{%
\begin{tabular}{c| 
    >{\columncolor[HTML]{FFF7E6}}c >{\columncolor[HTML]{FFF7E6}}c >{\columncolor[HTML]{FFF7E6}}c >{\columncolor[HTML]{FFF7E6}}c 
    >{\columncolor[HTML]{E6F2FF}}c >{\columncolor[HTML]{E6F2FF}}c >{\columncolor[HTML]{E6F2FF}}c >{\columncolor[HTML]{E6F2FF}}c 
    }
\toprule
& \multicolumn{4}{c}{\cellcolor[HTML]{FFF7E6}\textbf{GINE}} & \multicolumn{4}{c}{\cellcolor[HTML]{E6F2FF}\textbf{ResGatedGCN}} \\
    \cmidrule(lr){2-5} \cmidrule(lr){6-9}
    \textbf{\#layers} & \textbf{MAE$\downarrow$} & \textbf{RMSE$\downarrow$} & \textbf{R$^2\uparrow$} & \textbf{\#Param.} &
    \textbf{MAE$\downarrow$} & \textbf{RMSE$\downarrow$} & \textbf{R$^2\uparrow$} & \textbf{\#Param.} \\
    \midrule
    \textbf{1} & nan & nan & nan & 0.90M & nan & nan & nan & 1.95M \\
    \textbf{2} & 0.4451 & 0.5937 & 0.6975 & 1.03M & 0.4670 & 0.6197 & 0.6704 & 2.08M \\
    \textbf{3} & \textbf{0.4337} & \textbf{0.5804} & \textbf{0.7109} & 1.10M & 0.4576 & 0.6043 & 0.6866 & 2.15M \\
    \textbf{4} & 0.4359 & 0.5851 & 0.7062 & 1.16M & \textbf{0.4418} & \textbf{0.5922} & \textbf{0.6990} & 2.21M \\
    \bottomrule
  \end{tabular}%
  }
\end{table}

\begin{table}[h!]         
  \centering
  \caption{Routing on view~(d) vs.\ head layers (1--4).}
  \label{tab:route_d_headlayers_test}
  \resizebox{\linewidth}{!}{%
\begin{tabular}{c| 
    >{\columncolor[HTML]{FFF7E6}}c >{\columncolor[HTML]{FFF7E6}}c >{\columncolor[HTML]{FFF7E6}}c >{\columncolor[HTML]{FFF7E6}}c 
    >{\columncolor[HTML]{E6F2FF}}c >{\columncolor[HTML]{E6F2FF}}c >{\columncolor[HTML]{E6F2FF}}c >{\columncolor[HTML]{E6F2FF}}c 
    }
\toprule
& \multicolumn{4}{c}{\cellcolor[HTML]{FFF7E6}\textbf{GINE}} & \multicolumn{4}{c}{\cellcolor[HTML]{E6F2FF}\textbf{ResGatedGCN}} \\
    \cmidrule(lr){2-5} \cmidrule(lr){6-9}
    \textbf{\#layers} & \textbf{MAE$\downarrow$} & \textbf{RMSE$\downarrow$} & \textbf{R$^2\uparrow$} & \textbf{\#Param.} &
    \textbf{MAE$\downarrow$} & \textbf{RMSE$\downarrow$} & \textbf{R$^2\uparrow$} & \textbf{\#Param.} \\
    \midrule
    \textbf{1} & nan & nan & nan & 0.90M & nan & nan & nan & 1.95M \\
    \textbf{2} & 0.6111 & 0.792 & 0.5122 & 1.03M & 0.6007 & 0.7826 & 0.5237 & 2.08M \\
    \textbf{3} & \textbf{0.4637} & \textbf{0.6148} & \textbf{0.7060} & 1.09M & 0.5678 & 0.7551 & 0.5565 & 2.14M \\
    \textbf{4} & 0.4971 & 0.6636 & 0.6581 & 1.16M & \textbf{0.5445} & \textbf{0.7099} & \textbf{0.6089} & 2.21M \\
    \bottomrule
  \end{tabular}%
  }
\end{table}
The following four tables (\autoref{tab:place_d_gnnlayers}-\autoref{tab:route_d_headlayers_test}) present supplementary experiments for the Placement and Routing tasks under view (d). These experiments investigate the impact of two key architectural hyperparameters—the depth of the Graph Neural Network (GNN) backbone and the number of prediction head layers—on model performance. The results indicate that the number of GNN layers has a relatively minor and inconsistent effect, whereas the number of head layers is a crucial factor significantly influencing the outcomes.

\textbf{Impact of GNN Depth} (\autoref{tab:place_d_gnnlayers} and \autoref{tab:route_d_gnnlayers_test}):
The number of GNN layers (ranging from 3 to 6) does not exhibit a consistent impact on model performance. In \autoref{tab:place_d_gnnlayers} (Placement), GINE achieves its best results at deeper configurations, whereas ResGatedGCN performs best with fewer layers. A similar observation holds for routing in \autoref{tab:route_d_gnnlayers_test}, where both models achieve their strongest results at intermediate depths rather than at the deepest settings. Across both tasks, the performance metrics fluctuate across layer configurations without showing a clear monotonic trend. These results suggest that increasing GNN depth alone does not consistently improve prediction quality, and that message-passing depth is not a dominant factor for these tasks.

\textbf{Impact of Head Layers} (\autoref{tab:place_d_headlayers_test} and \autoref{tab:route_d_headlayers_test}):
In contrast, the number of head layers (ranging from 1 to 4) has a much stronger influence on performance. In \autoref{tab:place_d_headlayers_test} (Placement), increasing the depth of the prediction head leads to noticeable improvements for both GINE and ResGatedGCN. A similar trend is observed in \autoref{tab:route_d_headlayers_test} (Routing), where deeper prediction heads consistently produce better results. These improvements indicate that the design of the prediction head plays a crucial role in model performance. In particular, richer head architectures appear to provide stronger capacity for mapping learned node or edge embeddings to task-specific targets, highlighting the importance of prediction-head design in GNN-based physical design tasks.

\end{document}